\documentclass{article}
\usepackage{natbib,amsmath,graphics,epsf,jmlr2e}

\newcommand{\Lmax}   { {L_{\mathrm{max}}}}
\newcommand{\Lmin}{\Lambda}

\newcommand{\Past}	{ \stackrel{\leftarrow} {S} }
\newcommand{\past}	{ {\stackrel{\leftarrow} {s}} }

\newcommand{\Future}	{ \stackrel{\rightarrow}{S} }
\newcommand{\future}	{ \stackrel{\rightarrow}{s} }

\newcommand{\pastL}	{ {\stackrel{\leftarrow} {s}}^L }
\newcommand{\PastL}	{ {\stackrel{\leftarrow} {S}}^L }

\newcommand{\futureL}	{ {\stackrel{\rightarrow}{s}}^L }
\newcommand{\FutureL}	{ {\stackrel{\rightarrow}{S}}^L }

\newcommand{\AllPasts}	{ { \stackrel{\leftarrow} {\rm {\bf S}} } }

\newcommand{\CausalState}	{ {\cal S} }
\newcommand{\CausalStatePrime}	{ {\CausalState}^{\prime}}
\newcommand{\causalstate}	{ \sigma }
\newcommand{\CausalStateSet}	{ \boldsymbol{\CausalState} }
\newcommand{\AlternateState}	{ {\cal R} }
\newcommand{\alternatestate}	{ \rho }
\newcommand{\AlternateStateSet}	{ \boldsymbol{\AlternateState} }
\newcommand{\PrescientState}	{ \widehat{\AlternateState} }

\newcommand{\PrescientStateSet}	{ \boldsymbol{\PrescientState}}

\newcommand{\NextObservable}	{ {\stackrel{\rightarrow} {S}}^1 }

\newcommand{\Prob}		{ {\rm P}}

\newcommand{\LLimit}	{ {L \rightarrow \infty}}
\newcommand{\Cmu}		{ {C_\mu}}
\newcommand{\hmu}		{ {h_\mu}}

\newcommand{\Measurable}{ {\bf \mu}}

\newcommand{\EstCausalState}	{\widehat{\mathcal S}}
\newcommand{\estcausalstate}	{\widehat{\sigma}}
\newcommand{\EstCausalStateSet}	{\boldsymbol{\EstCausalState}}
\newcommand{\EstCausalFunc}	{\widehat{\epsilon}}

\newcommand{\ProcessAlphabet}	{\mathcal{A}}
\newcommand{\ProbEst}			{ {\widehat{\Prob}_N}}

\begin{document}

\jmlrheading{?}{2003}{?--?}{11/02}{?/?}{Cosma Rohilla Shalizi, Kristina Lisa Shalizi, and James P. Crutchfield}
\ShortHeadings{Pattern Discovery in Time Series}{Shalizi, Shalizi, and Crutchfield}
\firstpageno{1}

\title{An Algorithm for Pattern Discovery in Time Series}
\author{\name Cosma Rohilla Shalizi \email shalizi@santafe.edu \\
\addr Santa Fe Institute, 1399 Hyde Park Road, Santa Fe, NM 87501\\
\addr Center for the Study of Complex Systems, University of Michigan, Ann Arbor, MI 48109
\AND
\name Kristina Lisa Shalizi \email klinkner@santafe.edu \\
\addr Physics Department, University of San Francisco, 2130 Fulton Street, San
Francisco, CA 94117\\
\addr Santa Fe Institute, 1399 Hyde Park Road, Santa Fe, NM 87501\\
\addr Statistics Department, University of Michigan, Ann Arbor, MI 48109
\AND
\name James P. Crutchfield \email chaos@santafe.edu \\
\addr Santa Fe Institute, 1399 Hyde Park Road, Santa Fe, NM 87501}
\editor{}
\maketitle

\begin{abstract}

We present a new algorithm for discovering patterns in time series and other
sequential data.  We exhibit a reliable procedure for building the minimal set
of hidden, Markovian states that is statistically capable of producing the
behavior exhibited in the data --- the underlying process's \emph{causal
  states}.  Unlike conventional methods for fitting hidden Markov models (HMMs)
to data, our algorithm makes no assumptions about the process's causal
architecture (the number of hidden states and their transition structure), but
rather infers it from the data.  It starts with assumptions of minimal
structure and introduces complexity only when the data demand it.  Moreover,
the causal states it infers have important predictive optimality properties
that conventional HMM states lack.  We introduce the algorithm, review the
theory behind it, prove its asymptotic reliability, use large deviation theory
to estimate its rate of convergence, and compare it to other algorithms which
also construct HMMs from data.  We also illustrate its behavior on an example
process, and report selected numerical results from an implementation.

\end{abstract}

\begin{keywords}
Pattern discovery, hidden Markov models, variable length Markov models, causal
inference, statistical complexity, computational mechanics, information theory,
time series
\end{keywords}

\section{Introduction}

Recent years have seen a rising interest in the problem of \emph{pattern
discovery} \citep{Calculi-of-emergence,Hand-Mannila-Smyth,Spirtes-Glymour-Scheines}: Given data produced by a process, how
can one extract meaningful, predictive patterns from it, without fixing in
advance the kind of patterns one looks for? The problem represents the
convergence of several fields: unsupervised learning
\citep{Neural-Computation-unsupervised}, data mining and knowledge discovery
\citep{Hastie-Tibshirani-Friedman}, causal inference in statistics
\citep{Pearl-causality}, and the statistical physics of complex and
highly organized forms of matter
\citep{Badii-Politi,DNCO,CMPPSS,Varn-beyond-the-fault}.  It should be carefully
distinguished both from pattern recognition (essentially a matter of learning
to classify instances into known categories) and from merely building
predictors which, notoriously, may not lead to any genuine understanding of the
process.  We do not want to merely recognize patterns; we want to find the
patterns that are there to be recognized.  We do not only want to forecast; we
want to know the hidden mechanisms that make the forecasts work.  The point, in
other words, is to find the causal, dynamical structure intrinsic to the
process we are investigating, ideally to extract all the patterns in it that
have any predictive power.

Naturally enough, an account of what ``pattern'' means (especially in the
sequential setting) is crucial to progress in the above fields.  We and
co-authors have, over several years, elaborated such an account in a series of
publications on the \emph{computational mechanics} of dynamical systems
\citep{Inferring-stat-compl,Calculi-of-emergence,Upper-thesis,CMPPSS}.  Here,
we cash in these foundational studies to devise a new and practical algorithm
--- Causal-State Splitting Reconstruction (CSSR) --- for the problem of
predicting time series and other sequential data.  But, to repeat, we are not
giving yet another prediction algorithm; not only is the predictor delivered by
our algorithm provably optimal, in a straightforward sense that will be made
plain below, but it is also a \textit{causal} model.

In this paper, we first review the basic ideas and results of computational
mechanics, limiting ourselves to those we will use more or less directly.  With
this background in place, we present the CSSR algorithm in considerable detail.
We analyze its run-time complexity, prove its convergence using large-deviation
arguments, and discuss how to bound the rate of convergence.  By way of
elucidating CSSR we compare it to earlier computational mechanics algorithms
for causal-state reconstruction, and to the more familiar ``context-tree''
algorithms for inferring Markovian structure in sequences.  Our conclusion
summarizes our results and marks out directions for future theoretical work.

\section{Computational Mechanics}

Consider a discrete-time, discrete-valued stochastic process, $\ldots S_{-2}
S_{-1} S_0 S_1 S_2 \ldots$, with $S_i$ taking values in $\ProcessAlphabet$, a
finite alphabet of $k$ symbols.  At any time $t$, we can break the sequence of
random variables into a past (or history) $\Past_t$ and a future $\Future_t$,
both of which in general extend infinitely far.  Assume the process is
conditionally stationary; i.e., for all measurable future events $A$,
$\Prob(\Future_t \in A|\Past_t = \past)$ does not depend on $t$.\footnote{Every
stationary process is conditionally stationary.}  We accordingly drop the
subscript and simply speak of $\Past$ and $\Future$.  When we want to refer
only to the first $L$ symbols of $\Future$ we write $\FutureL$ and, similarly,
$\PastL$ denotes the last $L$ symbols of $\Past$.  (Lower-case versions of
these random variables denote particular instances.)

The problem of predicting the future is to go from $\Past$ to a guess at
$\Future$ or, more generally, a guess at the future distribution
$\Prob(\Future)$.  There is thus a function $\eta$ from particular histories
$\past$ to predictions $\Prob(\Future|\past)$.  Every prediction method
therefore imposes a partition on the set $\AllPasts$ of histories.  The cells
of this partition collect all the histories for which the same prediction is
made; i.e., $\past_1$ and $\past_2$ are in the same partition cell if and only
if $\eta(\past_1) = \eta(\past_2)$.  The cells of the partition are the
\emph{effective states} of the process, under the given predictor.  We will, by
a slight abuse of notation, also use $\eta$ as the name of the function from
histories to effective states.  The random variable for the current effective
state will be $\AlternateState$ and a particular value of it $\alternatestate$.
The set of states induced by $\eta$ will be denoted
$\AlternateStateSet$.\footnote{It is important to distinguish, here, between
  the effective states of a given predictor and alternative states used in
  particular representations of that predictor.  For general hidden Markov
  models, the hidden states are \emph{not} the same as the effective
  states of the predictor, since the former are not equivalence classes of
  histories.  It is always possible, however, to derive the effective states
  from the HMM states \citep{Upper-thesis}.}

Clearly, given a partition $\AlternateStateSet$ of $\AllPasts$, the best
prediction to make is whatever the distribution of futures, conditional on that
partition, happens to be: $\Prob(\Future|\AlternateState = \eta(\past))$.
Getting this conditional distribution right is nontrivial, but nonetheless
secondary to getting the partition right.  The result is that the problem of
finding an optimal predictor reduces (largely) to that of finding an optimal
partition of $\AllPasts$.

In what sense is one partition, one predictor, better than another?  One
obvious sense, inspired by information theory,\footnote{For brief definitions
of information-theoretic terms and notation, see the appendix.  For more
details, see \citet{Cover-and-Thomas}, whose notation we follow.} employs
the mutual information $I[\Future;\AlternateState]$ between the process's
future $\Future$ and the effective states $\AlternateStateSet$ to quantify
``goodness''. This quantity is not
necessarily well defined if we consider the entire semi-infinite future
$\Future$, since $I$ could be infinite.  But it is always well defined if we
look at futures $\FutureL$ of finite length.  Since
$I[\FutureL;\AlternateState] = H[\FutureL] - H[\FutureL|\AlternateState]$ and
the first term is the same for all sets of states, maximizing the mutual
information is the same as minimizing the conditional entropy.  There is a
lower bound on this conditional entropy, namely $H[\FutureL|\AlternateState]
\geq H[\FutureL|\Past]$ --- one can do no better than to remember the whole
past.  Call a set of states (a partition) that attains this lower bound for all
$L$ \emph{prescient}, and one which attains
it only for $L=1$ \emph{weakly prescient}.  We have shown elsewhere
\citep{CMPPSS} that prescient states are sufficient statistics for the
process's future and so, for any reasonable loss function, the optimal
prediction strategy can be implemented using prescient states
\citep{Blackwell-Girshick}.  We will therefore focus exclusively on conditional
entropy.

In general, there are many alternative sets of prescient states.  To select
among them, one can invoke Occam's Razor and chose the simplest set of
prescient states. (In the computational mechanics framework minimality is not
assumed --- it is a consequence; see below.) Complexity here is also measured
information-theoretically, by the Shannon entropy $H[\AlternateState]$ of the
effective states $\AlternateState$. This is the amount of information one needs
to retain about the process's history for prediction --- the amount of memory
the process retains about its own history.  For this reason, it is called the
\emph{statistical complexity}, also written $\Cmu(\AlternateState)$.  Restated
then, our goal is to find a set of prescient states of minimal statistical
complexity.

It turns out that, for each process, there is a unique set of prescient states
that minimizes the statistical complexity.  These are the \emph{causal states}.

\begin{definition}[A Process's Causal States]
The \emph{causal states} of a process are the members of the range of a
function $\epsilon$ that maps from histories to sets of histories.  If
$\Measurable(\Future)$ is the collection of all measurable future events, then
\begin{eqnarray}
\epsilon (\past) & \equiv \{ \past^\prime \in \AllPasts & | \forall F \in
  \Measurable(\Future), ~\Prob(\Future \in F | \Past = \past ) = \Prob(\Future
  \in F | \Past = \past^\prime )\} ~,
\label{def-of-causal-states}
\end{eqnarray}
Write the $i^{th}$ causal state as $\causalstate_i$ and the set of all causal
states as $\CausalStateSet$; the corresponding random variable is denoted
$\CausalState$ and its realization $\causalstate$.
\label{CausalStatesFunctionDefn}
\end{definition}

A consequence of this definition is that, for any $L$,
\begin{eqnarray}
\epsilon(\past) & = & \{ \past^\prime
  | \Prob(\FutureL  =  \futureL | \Past = \past)
  = \Prob(\FutureL = \futureL | \Past = \past^\prime ) ~, 
\forall  \futureL \in \FutureL, \past^\prime \in \Past \} ~.
\label{finite-def-of-causal-states}
\end{eqnarray}

Given any other set of prescient states, say $\PrescientStateSet$, there is a
function $f$ such that $\CausalState = f(\PrescientState)$ almost always.
Hence the entropy of the causal states --- their statistical complexity --- is
less than or equal to that of the prescient states.\footnote{Notably,
the equivalence-class definition of causal states leads directly
to their being the minimal set. That is, one does {\em not} have
to invoke Occam's Razor, it is derived as a consequence of the
states being causal.} Moreover, if
$\Cmu(\PrescientState) = \Cmu(\CausalState)$, then $f$ is invertible and the
two sets of states are equivalent, up to relabeling (and a possible set of
exceptional histories of measure zero).  Statistically speaking, the
causal states are the minimal sufficient statistic for predicting the process's
future.

Given an initial causal state and the next symbol from the original process,
only certain successor causal states are possible.  Thus, we may define allowed
transitions between causal states and the probabilities of these transitions.
Specifically, the probability of moving from state $\causalstate_i$ to state $\causalstate_j$ on
symbol $s$ is
\begin{equation}
{\rm T}^{(s)}_{ij}
  \equiv \Prob(\Future^1 = s, \CausalStatePrime = \causalstate_j
  | \CausalState = \causalstate_i) ~.
\end{equation}
Note that
\begin{eqnarray*}
\sum_{s \in \ProcessAlphabet}{\sum_{\causalstate_j \in \CausalStateSet}
  {{\rm T}^{(s)}_{ij}}}
  & = & \sum_{s\in\ProcessAlphabet}{\Prob(\Future^1 = s
  | \CausalState = \causalstate_i)} \\
  & = & 1~.
\end{eqnarray*}
We denote the set of these \emph{labeled transition probabilities} by
$\mathbf{T} = \{ {\mathrm T}^{(s)}_{ij} : ~ s \in \ProcessAlphabet,
~ \causalstate_i, \causalstate_j \in \CausalStateSet \}$.  The combination of
states and transitions is called the
process's \emph{$\epsilon$-machine}; it represents the way in which the process
stores and transforms information \citep{Inferring-stat-compl}.  Examples for a
number of different kinds of process --- including HMMs, cellular automata, and
chaotic dynamical systems --- are given in \citet{Calculi-of-emergence};
for applications to empirical data, see \citet{Palmer-complexity-in-atmo}
and \citet{Watkins-et-al-comp-mech-of-geomag}.  The uncertainty in the
next symbol given the current state, $H[\NextObservable|\CausalState]$, is
exactly the process's entropy rate $\hmu$.

\subsection{Properties of Causal States and $\epsilon$-Machines}

Here we state the main properties of causal states which we use below.  See
\citet{CMPPSS} for proofs.

\begin{proposition} 
The causal states are \emph{homogeneous} for future events: That is, all
histories belonging to a single causal state $\causalstate$ have the same
conditional distribution $\Prob(\Future|\causalstate)$ of future events.  The
causal states are also the largest sets (partition elements) that are
homogeneous for future events: every history with that future conditional
distribution is in that state.
\end{proposition}

\begin{proposition}
The process's future and past are independent given the causal state.
\label{prop:cond-ind-of-past-and-future}
\end{proposition}

\begin{proposition}
The causal states themselves form a Markov process.
\label{prop:causal-states-are-Markov}
\end{proposition}

\begin{proposition}
The $\epsilon$-machine is \emph{deterministic} in the sense of automata
theory; i.e., given the current state and the next symbol, there is a unique
successor state.
\label{prop:causal-states-are-deterministic}
\end{proposition}

Following practice in symbolic dynamics (see, e.g., \citet{Lind-Marcus}), we
also say that the causal states are \emph{future resolving}. That is to say,
for each state $\causalstate_i$ and symbol $s$, there is at most one state
$\causalstate_j$ such that ${\rm T}^{(s)}_{ij} > 0$.

\begin{proposition}
Any prescient set $\PrescientStateSet$ of states is a refinement of the causal
states $\CausalStateSet$.
\label{prop:prescient-refinement}
\end{proposition}

For more on these properties, and the reasons for calling $\epsilon$-machine
states \textit{causal}, see \citet{CMPPSS}, especially Sections II--IV.

Finally, there is one additional result which is so crucial to CSSR that we
give it together with a summary of its proof.

\begin{proposition}
If a set of states is weakly prescient and deterministic, then it is prescient.
\label{prop:homogeneous-resolving}
\end{proposition}

\emph{Proof}: A rough way to see this is to imagine using weak prescience to
obtain a prediction for the next symbol, updating the current state using
determinism, then getting a prediction for the second symbol from weak
prescience again, and so on, as far forward as needed.

More formally, we proceed by induction.  Suppose a deterministic set
$\AlternateStateSet$ of states has the same distribution for futures of length
$L$ as do the causal states.  This implies that they have the same distribution
for futures of shorter lengths, in particular for futures of length 1.  We can
use that fact, in turn, to show that they must have the distribution for
futures of length $L+1$:
\begin{eqnarray*}
\Prob(\Future^{L+1} = s^L a|\AlternateState = \eta(\past))
  & = & \Prob(\Future_{L+1} = a|\AlternateState = \eta(\past), \Future^L = s^L)
	\Prob(\Future^L = s^L|\AlternateState = \eta(\past))\\
  & = & \Prob(\Future_{L+1} = a|\AlternateState = \eta(\past), \Future^L = s^L)
	\Prob(\Future^L = s^L|\CausalState = \epsilon(\past)) ~.
\end{eqnarray*}
Determinism implies
\begin{equation}
\nonumber
\Prob(\Future_{L+1} = a|\AlternateState = \eta(\past), \Future^L = s^L)
  = \Prob(\Future^1 = a|\AlternateState = \eta(\past s^L)) ~.
\end{equation}
So
\begin{eqnarray*}
\Prob(\Future^{L+1} = s^L a|\AlternateState = \eta(\past)) & = &
\Prob(\Future^1 = a|\AlternateState = \eta(\past s^L)) \Prob(\Future^L =
s^L|\CausalState = \epsilon(\past))\\
& = & \Prob(\Future^1 = a|\CausalState = \epsilon(\past s^L)) \Prob(\Future^L = s^L|\CausalState = \epsilon(\past))\\
& = & \Prob(\Future^{L+1} = s^L a|\CausalState = \epsilon(\past)) ~.
\end{eqnarray*}

$\Box$

Since the causal states fulfill the hypotheses of the proposition and are
minimal among all prescient sets of states, they are also minimal among all
sets of states that are deterministic and weakly prescient.

\subsection{Recurrent, Transient, and Synchronization States}

Proposition \ref{prop:causal-states-are-Markov} tells us that the causal states
form a Markov process.  The states are therefore either \emph{recurrent}, i.e.,
returned to infinitely often, or \emph{transient}, visited only finitely often
with positive probability \citep{Grimmett-Stirzaker}.  For us, the recurrent
states represent the actual causal structure of the process and, as such, they
are what we are truly interested in \citep{Upper-thesis}.  The most important
class of transient states, and indeed the only ones encountered in practice,
are the \emph{synchronization states}, which can never be returned to, once a
recurrent state has been visited.  The synchronization states represent
observational histories that are insufficient to fix the process in a definite
recurrent state.  Given the recurrent states, there is a straightforward
algorithm \citep{DNCO} which finds the synchronization states. Here, however,
we prefer to omit them from our algorithm's final results altogether. While the
algorithm will find them, it will also prune them, reporting only the truly
structural, recurrent states.

For a complete taxonomy of causal states, including a discussion of issues of
synchronization and reachability, see \citet{Upper-thesis}.

\section{Causal-State Splitting Reconstruction}

Here we introduce the Causal-State Splitting Reconstruction (CSSR) algorithm
that estimates an $\epsilon$-machine from samples of a process.  The algorithm
is designed to respect the essential properties of causal states just
outlined.  

The basic idea of CSSR is straightforward and similar to
state-splitting methods for finite-state machines \citep{Lind-Marcus}.  It
starts out assuming a simple model for the process and elaborates model
components (adds states and transitions) only when statistically justified.
More specifically, CSSR begins assuming the process is independent, identically
distributed (IID) over the alphabet $\ProcessAlphabet$.  This is equivalent to
assuming the process is structurally simple and is as random as possible.  One
can work through Definition \ref{CausalStatesFunctionDefn} given above for
causal states to show that an IID process has a single causal state.  Thus,
initially the process is seen as having zero statistical complexity ($\Cmu =
H[\CausalState] = \log_2 1 = 0$) and high entropy rate $\hmu \leq \log_2 k$.
From this starting point, CSSR uses statistical tests to see when it must add
states to the model, which increases the estimated complexity, while lowering
the estimated entropy rate.  The initial model is kept only if the process
actually is IID.

A key and distinguishing property of CSSR is that it maintains homogeneity of
the causal states and determinism of the state-to-state transitions as the
model grows.  The result is that at each stage the estimated model is an
$\epsilon$-machine, satisfying the criteria above, for an approximation of the
process being modeled.  One important consequence is that the degree of
unpredictability (the process's entropy rate $\hmu$) can be directly estimated
from the approximate $\epsilon$-machine.\footnote{In general this calculation
\emph{cannot} be done directly for standard HMMs, which are nondeterministic
\citep{Blackwell-identifiability}.}

\subsection{The Algorithm}

Assume we are given a sequence of length $N$ over the finite alphabet
$\ProcessAlphabet$.  We wish to estimate from this a class $\EstCausalStateSet$
of effective states.  Each member $\estcausalstate$ of $\EstCausalStateSet$ is
a set of histories.  Say that a string $w$ is a \emph{suffix} of the history
$\past$ if $\pastL = w$ for some $L$, i.e., if the end of the history matches
the string.  To emphasize that a given string is a suffix, we write it as $*w$;
e.g., $*01011$.  We represent a state $\estcausalstate$ as a set of suffixes.
The function $\EstCausalFunc$ maps a finite history $\past$ to that
$\estcausalstate$ which contains a suffix of $\past$. We shall arrange it
so that the assignment of histories to states is never ambiguous.

One suffix $*w$ is the \emph{child} of another $*v$, if $w = av$, where $a$
is a single symbol.  That is, a child suffix is \emph{longer}, going into the past, by
one symbol than its \emph{parent}.  A suffix $*w$ is a \emph{descendant} of
its \emph{ancestor} $*v$ if $w = uv$, where $u$ is any (non-null) string.

In addition to a set of suffixes, each $\estcausalstate \in \EstCausalStateSet$
is associated with a distribution for the next observable $\NextObservable$;
i.e., $\Prob(\NextObservable = a|\EstCausalState = \estcausalstate)$ is defined
for each $a \in \ProcessAlphabet$ and each $\estcausalstate$.  We call this
conditional distribution the state's {\em morph}.

The \emph{null hypothesis} is that the process being modeled is Markovian on the basis
of the states in $\EstCausalStateSet$; that is,
\begin{eqnarray}
\Prob({\Future}^{1}|{\Past}^{L} = a {s}^{L-1}) & = &
\Prob({\Future}^{1}|\EstCausalState = \EstCausalFunc(s^{L-1})) ~.
\end{eqnarray}

Naturally, one can apply a standard statistical test --- e.g., the
Kolmogorov-Smirnov (KS) test\footnote{See \citet[sec.\
14.3]{Numerical-Recipes-in-C} and \citet[pp.\
178--187]{Hollander-Wolfe-nonparametric-methods} for details of this test.} ---
to this hypothesis at a specified significance level, denoted $\alpha$.  If one
uses the KS test, as we do here, one avoids directly estimating the morph
conditional distribution and simply uses empirical frequency counts.  Recall
that the significance level is the probability of type-I error (rejecting the
null when it is true).  Generally, the KS test has higher power than other,
similar tests, such as ${\chi}^2$ \citep{Rayner-Best-smooth-tests}.  That is,
the KS test has a lower probability of type-II error, of accepting the null
hypothesis when it is false.  Empirically, the precise test we use makes
little difference to the algorithm.
 
We modify the set $\EstCausalStateSet$ only when the null hypothesis is
rejected.  When we reject the null hypothesis, we fall back on a
\emph{restricted alternative hypothesis}, which is that we have the right set
of conditional distributions, but have assigned them to the wrong histories.
We therefore try to assign child suffixes whose morphs differ from their
parents to existing states.  Only if this alternative is itself rejected do we
create additional, new distributions to explain or capture the apparent
non-Markovianness.  This increases the cardinality of $\EstCausalStateSet$.

Throughout, we shall write $\nu(A)$ for the number of times the event $A$
happens in the data $s^N$ --- the \textit{count} of $A$.

Thus, there are four CSSR parameters: the measurement alphabet size $k$, the
length $N$ of the data stream $s^N$, the length $\Lmax$ of the longest history
to be considered, and the significance level $\alpha$ of the null-hypothesis
statistical test.  There are three procedures in CSSR: Initialize, Homogenize,
and Determinize.

\vspace{5mm}

\noindent \textbf{I.~Initialize} Set $L=0$ and $\EstCausalStateSet =
\left\{\estcausalstate_0\right\}$, where $\estcausalstate_0 =
\left\{* \lambda\right\}$; i.e., $\estcausalstate_0$ contains only the null
sequence $\lambda$.  We regard $* \lambda$ as a suffix of any history, so that
initially $\EstCausalFunc$ maps all histories to $\estcausalstate_0$.  The
morph of $\estcausalstate_0$ is defined by
\begin{equation}
  \Prob(\NextObservable = a|\EstCausalState = \estcausalstate_0)
  = \Prob(\NextObservable = a) ~,
  \label{IIDBaseCase}
\end{equation}
so the initial model is that the process is a sequence of independent,
identically-distributed random variables.  As a consequence, the statistical
complexity vanishes ($\Cmu(\EstCausalStateSet) = \log_2 1 = 0$) and the
entropy rate is maximal
($\hmu(\EstCausalStateSet) = H[\NextObservable = a] \leq \log_2 k$).

\vspace{5mm}

\noindent \textbf{II.~Homogenize} We first generate states whose members are
homogeneous for the next symbol --- states whose histories all lead to the same
morph.  Said differently, we generate states whose member histories have no
\textit{significant} differences in their individual morphs.  We do this
as follows.

\begin{enumerate}
\setlength{\itemsep}{0mm}
\item For each $\estcausalstate \in \EstCausalStateSet$, calculate
  $\ProbEst(\NextObservable|\EstCausalState = \estcausalstate)$ --- the
  future distribution from that state, given the data sequence.
  \begin{enumerate}
  \setlength{\topsep}{0mm}
  \setlength{\itemsep}{0mm} 
  \item When $L=0$ and the only suffix is $*\lambda$ and we could have seen any
	history; so we use Eq. (\ref{IIDBaseCase}) above.
  \item For each sequence $\pastL \in \estcausalstate$, estimate
    $\ProbEst(\NextObservable = a|\PastL = \pastL)$.  The naive
	maximum-likelihood estimate,
    \begin{equation}
      \ProbEst(\NextObservable = a|\PastL = \pastL)
      = \frac{\nu(\PastL = \pastL, \NextObservable = a)}{\nu(\PastL = \pastL)} ~,
      \label{GivenHistoryMorphEst}
    \end{equation}
  is simple and well adapted to the later parts of the procedure, but other
  estimators could be used.  This distribution is the morph of the history
  $\pastL$.
  \item The morph of the state $\estcausalstate$ is the weighted average of the
    morphs of its histories $\pastL \in \estcausalstate$, with weights
    proportional to $\nu(\PastL = \pastL)$:
  \begin{equation}
  \ProbEst(\NextObservable = a | \estcausalstate)
	 = \frac{1}{z} \sum_{\pastL \in \estcausalstate} 
	 \nu(\PastL = \pastL) \ProbEst(\NextObservable = a|\PastL = \pastL) ~,
  \end{equation}
    where $z = \sum_{\pastL \in \estcausalstate}{\nu(\PastL = \pastL)}$ is the
    number of occurrences in $s^N$ of suffixes in $\estcausalstate$.
  \end{enumerate}
\item For each $\estcausalstate \in \EstCausalState$, test the null (Markovian)
  hypothesis.  For each length-$L$ history $\pastL \in \estcausalstate$ and
  each $a \in \ProcessAlphabet$, generate the suffix $a \pastL$ of length
  $L+1$ --- a {\em child suffix} of $\pastL$.
  \begin{enumerate}
  \setlength{\topsep}{0mm}
  \setlength{\itemsep}{0mm} 
  \item Estimate the morph of $a \pastL$ by the same method as used above, Eq.
    (\ref{GivenHistoryMorphEst}).
  \item If the morphs of $a \pastL$ and $\estcausalstate$ do not differ
    according to the significance test, add $a \pastL$ to $\estcausalstate$.
  \item If they do differ, then test whether there are any states in
    $\EstCausalStateSet$ whose morphs do \textit{not} differ significantly from
    that of $a \pastL$.  If so, add $a \pastL$ to the state whose morph its
    morph matches most closely, as measured by the score of the significance
    test.\footnote{Actually, to which of these states $a \pastL$ is assigned is
      irrelevant in the limit where $N \rightarrow \infty$; but the choice we
      use here is convenient, plausible, and can be implemented consistently.}
    (This is the ``restricted alternative hypothesis'' mentioned above.)
  \item However, if the morph of $a \pastL$ is significantly different from the
	morphs of all existing states, then create a new state and add $a \pastL$
	to it.
  \item Recalculate the morphs of states from which sequences have been
	  added or deleted.
  \end{enumerate}
\item Increment $L$ by one.
\item Repeat steps 1--3 until reaching the maximum history length $\Lmax$.
\end{enumerate}
\noindent
At the end of state homogenization, no history is in a state whose morph is
significantly different from its own.  Moreover, every state's morph is
significantly different from every other state's morph.  The causal states have
this property, but their transitions are also deterministic and so we need
another procedure to ``determinize'' $\EstCausalStateSet$ (see Proposition
\ref{prop:homogeneous-resolving}).

\vspace{5mm}

\noindent \textbf{III.~Determinize}
\begin{enumerate}
\setlength{\itemsep}{0mm} 
\item Eliminate transient states from the current state-transition structure,
  leaving only recurrent states.
\item For each state $\estcausalstate \in \EstCausalStateSet$:
  \begin{enumerate}
  \setlength{\itemsep}{0mm} 
  \item For each $a \in \ProcessAlphabet$:
	\begin{enumerate}
	\setlength{\itemsep}{0mm} 
	\item Calculate $\EstCausalFunc(\past a)$ for all $\past \in
	  \estcausalstate$ --- these are the {\em successor states on symbol
	    $a$} of the histories $\past$ --- by finding
		$\estcausalstate^\prime \in \EstCausalStateSet$ such that
		$(\pastL a)^L \in \estcausalstate^\prime$.
	\item If there are $n > 1$ successor states on $a$, create $n-1$ new states,
	  each with $\estcausalstate$'s ($L=1$) morph. Partition histories in
	  $\estcausalstate$ between $\estcausalstate$ and the new states so that
	  all histories in $\estcausalstate$ and the new states have the same
	  successor on $a$. Go to i.
	\end{enumerate}
  \item If every history $\past \in \estcausalstate$ has the same successor
    on $a$, for every $a$, go on to the next state.
  \end{enumerate}
\item From the new, deterministic states eliminate those which are transient.
\end{enumerate}

Since this procedure only produces smaller (fewer-suffix) states, and a state
with one suffix in it cannot be split, the procedure terminates, if only
by assigning each history its own state.  When it terminates,
$\EstCausalStateSet$ will be a set of states with deterministic transitions.
Moreover, since we create deterministic states by splitting homogeneous states,
the deterministic states remain homogeneous.

Now, as we noted, the causal states are the minimal states that have a
homogeneous distribution for the next symbol and are deterministic.  If we had
access to the exact conditional distributions from the underlying process,
therefore, and did not have to estimate the morphs, this procedure would return
the causal states.  Instead, it returns a set of states that, in a sense set
by the chosen significance test, cannot be statistically distinguished from
them.

We have implemented CSSR, and we report the results of numerical experiments on
an example problem in Section \ref{sec:dyn-learn} below.  The next
section outlines the example and illustrates CSSR's behavior.

\subsection{Example: The Even Process}
\label{sec:even-process-example}

To illustrate the workings of CSSR, let's see how it reconstructs an
easy-to-describe process with two states --- the even process of
\citet{Weiss-1973} --- that, despite its simplicity, has several nontrivial
properties. We use data from a simulation of the process and typical parameter
settings for the algorithm.

\begin{figure}
\begin{center}
\resizebox{2.0in}{1.33in}{\includegraphics{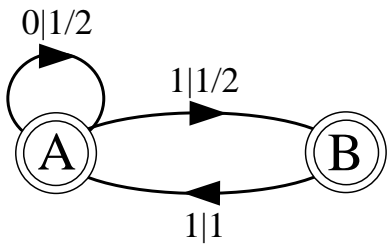}}
\caption{The even process: a strictly sofic, non-Markovian stochastic process.}
\label{figure:EvenProcess}
\end{center}
\end{figure}

We have a two-symbol alphabet, $\ProcessAlphabet = \left\{0, 1\right\}$. There
are two recurrent states, labeled $A$ and $B$ in Fig. \ref{figure:EvenProcess}.
State $A$ can either emit a $0$ and return to itself, or emit a $1$ and go to
$B$; we take the version where these options are equally likely. State $B$
always emits a $1$ and goes to $A$. The labeled transition matrices
$\mathbf{T}$ are thus
\begin{equation}
{\rm T}^{(0)}
  = \left[ \begin{matrix} 0.5 & 0 \cr 0 & 0 \cr \end{matrix} \right]
  {\rm ~~~and~~~}
{\rm T}^{(1)}
  = \left[ \begin{matrix} 0 & 0.5 \cr 1.0 & 0 \cr \end{matrix} \right] ~.
\end{equation}

We take $\Lmax = 3$ and $\alpha = 0.01$. We simulated the even process for
$10^4$ time steps and accumulated the sequence statistics for words of length
$4$ and shorter, given in Table \ref{EvenProcessCounts}. Since one slides a
window of size $\Lmax + 1 = 4$ through the data stream, there are $9,996$
length-$4$ words. Observe in Table~\ref{EvenProcessCounts} that sequences
containing odd-length blocks of $1$s, bounded by $0$s, do not appear. That is,
words of the form $\{ 01^{2k+1}0, ~k = 1, 2, \ldots \}$ are forbidden.

\begin{table}
\begin{center}
\begin{tabular}{lr||lr||lr||lr|lr}
Word & Count& Word	& Count	& Word	& Count & Word	& Count & Word	& Count\\
\hline
0	& 3309	& 00	& 1654	& 000	&  836 & 0000	& 414	& 1000	& 422\\
1	& 6687	& 01	& 1655	& 001	&  818 & 0001	& 422	& 1001	& 396\\
	&		& 10	& 1655	& 010	&    0 & 0010	&   0	& 1010	&   0\\
	&		& 11	& 5032	& 011	& 1655 & 0011	& 818	& 1011	& 837\\
	&		&		&		& 100	&  818 & 0100	&   0	& 1100	& 818\\
	&		&		&		& 101	&  837 & 0101	&   0	& 1101	& 836\\
	&		&		&		& 110	& 1654 & 0110	& 814	& 1110	& 841\\
	&		&		&		& 111	& 3378 & 0111	& 841	& 1111	& 2537
\end{tabular}
\caption{Count statistics for words of length $4$ and shorter from $10^4$
  samples of the even process of Fig. \ref{figure:EvenProcess}. The total
  count at each length is $9,996$ words.}
\label{EvenProcessCounts}
\end{center}
\end{table}

Overall, the data contains 3309 $0$s and 6687 $1$s, since for simplicity we fix
the total number of words at each length to be $9,996$. The initial state
$A_{L=0}$ formed at $L = 0$, containing the null suffix $*\lambda$, therefore
produces $1$s with probability $\approx 0.669$.

The null suffix $*\lambda$ has two children, $*0$ and $*1$. At $L=1$ the
probability of producing a $1$, conditional on the suffix $*0$, is
$\Prob(1|*0) \approx 1655/3309 \approx 0.500$, which is significantly different
from the distribution for the null suffix. Similarly, the probability of
producing a $1$, conditional on the suffix $*1$, is $\Prob(1|*1) \approx
5032/6690 \approx 0.752$, which is also significantly different from that of
the parent state. We thus produce two new states $B_{L=1}$ and $C_{L=1}$, one
containing the suffix $*1$ and one the suffix $*0$, respectively. There is now
a total of three states: $A_{L=0} = \{*\lambda\}$, $B_{L=1} = \{*1\}$, and
$C_{L=1} = \{*0\}$.

Examining the second generation of children at $L = 2$, one finds the
conditional probabilities of producing $1$s given in
Table~\ref{ConditionalCountsChildrenL_2}.
\begin{table}
\begin{center}
\begin{tabular}{ll}
Suffix $*w$ & $\Prob(1|*w)$ \\
\hline
*00 & $818/1654 \approx 0.495$\\
*10 & $837/1655 \approx 0.506$\\
*01 & $1655/1655 = 1.000$\\
*11 & $3378/5032 \approx 0.671$
\end{tabular}
\caption{Conditional counts having read two symbols.}
\label{ConditionalCountsChildrenL_2}
\end{center}
\end{table}
The morphs of the first two suffixes, $*00$ and $*10$, do not differ
significantly from that of their parent $*0$, so we add them to the parent
state $C_{L=1}$. But the second two suffixes, $*01$ and $*11$, must be split
off from their parent $B_{L=1}$. The morph for $*01$ does not match that of
$A_{L=0}$, so we create a new state $D_{L=2} = \{*01\}$. However, the morph
for $*11$ does match that of $A_{L=0}$ and so we add $*11$ to $A_{L=0}$.
At this point, there are four states: $A_{L=0} = \{*\lambda, *11\}$,
$B_{L=1} = \{*1\}$, $C_{L=1} = \{*0, *00, *10 \}$, and $D_{L=2} = \{*01\}$.

Let us examine the children of $C_{L=1}$'s $L = 2$ suffixes, whose statistics
are shown in Table~\ref{ConditionalCountsChildrenOfC1}.
\begin{table}
\begin{center}
\begin{tabular}{ll}
Suffix $*w$ & $\Prob(1|*w)$ \\
\hline
*000 & $422/836 \approx 0.505$ \\
*100 & $396/818 \approx 0.484$ \\
*010 & $0/837 = 0.0$ \\
*110 & $837/1655 \approx 0.506$
\end{tabular}
\caption{Conditional counts for child suffixes of state $C_{L=1}$'s
  $L=2$ suffixes.}
\label{ConditionalCountsChildrenOfC1}
\end{center}
\end{table}
None of these split and they are added to $C_{L=1}$.

The children of $D_{L=2}$ are given in
Table~\ref{ConditionalCountsChildrenOfD2}. Again, neither of these must be
split from $D_{L=2}$ and they are added to that state.
\begin{table}
\begin{center}
\begin{tabular}{ll}
Suffix *w & $\Prob(1|*w)$ \\
\hline
*001 & $818/818 = 1.000$ \\
*101 & $837/837 = 1.000$
\end{tabular}
\caption{Conditional counts for child suffixes of state $D_{L=2}$.}
\label{ConditionalCountsChildrenOfD2}
\end{center}
\end{table}

Finally, the children of $A_{L=0}$'s $L = 2$ suffix are given in
Table~\ref{ConditionalCountsChildrenOfA0}.
\begin{table}
\begin{center}
\begin{tabular}{ll}
Suffix *w & $\Prob(1|*w)$ \\
\hline
*011 & $841/1655 \approx 0.508$ \\
*111 & $2537/3378 \approx 0.751$
\end{tabular}
\caption{Conditional counts for child suffixes of state $A_{L=0}$'s $L=2$
  suffix $*11$.}
\label{ConditionalCountsChildrenOfA0}
\end{center}
\end{table}
$*011$ must be split off from $A_{L=0}$, but now we must check whether its morph
matches any existing distribution. In fact, its morph does not significantly
differ from the morph of $C_{L=1}$, so we add $*011$ to $C_{L=1}$. $*111$
must also be split off, but its morph is similar to $B_{L=1}$'s and so it is
added there. We are still left with four states, which now contain the
following suffixes: $A_{L=0} = \{*\lambda, *11\}$, $B_{L=1} = \{*1, *111\}$,
$C_{L=1} = \{*0, *00, *10, *000, *100, *110, *011\}$, and
$D_{L=2} = \{*01, *001, *101\}$.

\begin{figure}
\begin{center}
\resizebox{4.0in}{1.67in}{\includegraphics{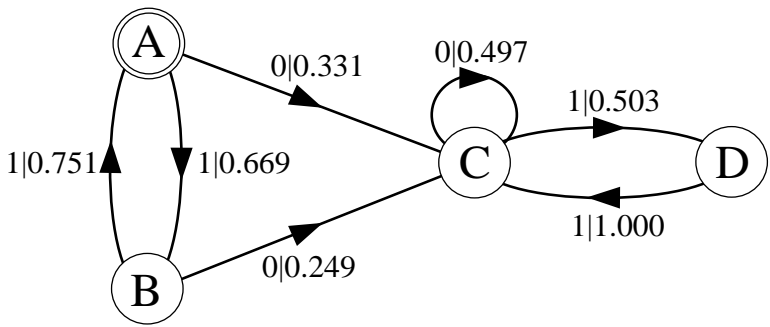}}
\end{center}
\caption{The full $\epsilon$-machine reconstructed from $10^4$ samples of the even process
  of Fig.\ \ref{figure:EvenProcess}. State $A$ (inscribed circle) is the
  unique start state.}
\label{figure:EvenProcessEMachine}
\end{figure}

Since we have come to the longest histories, $\Lmax = 3$, with which we
are working, Procedure II (Homogenize) is complete. We begin Procedure III
by checking which states have incoming transitions.  (We now drop the
time-of-creation subscript.)  State $C$ can be reached either from itself,
state $A$, or state $B$, all on a $0$. State $C$ is reached from $D$ on a
$1$ and state $D$ can be reached from state $C$ on a $1$. Finally, state $A$
can only be reached from $B$ on a $1$ and $B$ only on a $1$ from $A$. Figure
\ref{figure:EvenProcessEMachine} shows the resulting estimated full $\epsilon$-machine,
with transition probabilities.

Continuing on, though, we note that states $A$ and $B$ are transient.
Accordingly, we eliminate them and go on to check for determinism. Every suffix
in state $C$ goes to one in state $C$ by adding a $0$ and to one in state $D$
by adding a $1$, so we do not split $C$.  Similarly, every suffix in $D$ goes to
one in $C$ on adding a $1$, and state $D$ never produces $0$s.  Hence $D$ also
does not need to be changed.  Since both states are future-resolving, the
system of states is too, and we exit the determinization loop.  Finally, we
have not created any new states while determinizing, so we do not need to prune
new transient states.

The final result is an $\epsilon$-machine with the two states $C$ and $D$ and with the
estimated labeled transition probabilities
\begin{eqnarray*}
\nonumber
{\widehat{\rm T}}^{(0)}  = 
  \left [ \begin{matrix} 0.497 & 0 \cr 0 & 0 \cr \end{matrix} \right ]
&  ~{\rm and} ~ &
{\widehat{\rm T}}^{(1)}  = 
  \left [ \begin{matrix} 0 & 0.503 \cr 1.000 & 0 \cr \end{matrix} \right ]
  ~.
\end{eqnarray*}

We have just seen CSSR reconstruct the even process with length-$3$
histories.  Nonetheless, it is important to realize that the even process is
\textit{not} a third-order Markov chain. Since it matters whether the current
block of $1$s is of even or odd length, the process has a kind of infinite
memory. It is in fact \emph{strictly sofic} \citep{Weiss-1973}, meaning that,
while it has a finite number of states, it is not equivalent to a Markov chain
of any finite order. For this reason conventional Markov-model techniques find
it difficult to learn strictly sofic processes. In fact, many standard methods
cannot handle them at all. CSSR's ability to do so follows quite naturally from
its design, dictated in turn by the basic results of computational
mechanics.  (For more on these points, see Section \ref{sec:VLMMs}.)

\subsection{Time Complexity}
\label{sec:time-complexity}

Procedure I (Initialize) must compute the relative frequency of all words in
the data stream, up to length $\Lmax + 1$.  We can accomplish this in a single
pass through the data, storing the frequencies in a parse tree, with total time
proportional to the number $N$ of symbols in the data stream.  Thereafter we
never need to refer to the data again, just the parse tree.

Procedure II (Homogenize) checks, for each suffix, whether it belongs to the
same state as its parent.  This operation, along with repartitioning and the
creation of a new state, if needed, can all be done in constant time (through
use of a hash table).  Since there are at most $s(k,\Lmax) =
(k^{\Lmax+1}-1)/(k-1)$ suffixes, the total time for Procedure II is
proportional to $s(k,\Lmax)$.  Asymptotically, this is $O(k^\Lmax)$.

We can divide the time needed for Procedure III (Determinize) into three parts:
getting the transition structure, removing transient states, and determinizing.
(We remove transient states twice, but this only introduces a constant factor.)
The time to find the transition structure is at most $k s(k,\Lmax)$.
Removing transients can be done by finding the strongly connected components of
the state-transition graph, and then finding the recurrent part of the
component graph; both operations take a time proportional to the number of
nodes and edges in the state-transition graph.  The number of nodes can't be
more than $s(k,\Lmax)$, since there must be at least one suffix in each.
Similarly the number of edges can't be more than $k s(k,\Lmax)$ --- i.e., $k$
edges per suffix.  Hence transient-removal is $O(s(k,\Lmax)(k+1)) = O(k^{\Lmax
  + 1} + k^\Lmax) = O(k^{\Lmax+1})$.  The time needed to make one determinizing
pass is $k s(k,\Lmax)$, and the maximum number of passes needed is
$s(k,\Lmax)$, since this is the worst case for the number of states which could
be made. So the worst-case time for determinization is $O(k s^2(k,\Lmax)) =
O(k^{2\Lmax + 1})$.  Adding up for Procedure III, we get $O(k^{\Lmax + 1} +
k^{\Lmax + 1} + k^{2\Lmax + 1}) = O(k^{2\Lmax + 1})$.  Note that if removing
transients consumes the maximal amount of time, then determinization cannot and
vice versa.

Putting this together, we get $O(k^\Lmax) + O(k^{2\Lmax+1}) + O(N)$, which is
${O(k^{2\Lmax+1}) + O(N)}$ asymptotically. Observe that this is linear in
the data size $N$.  It is exponential in the alphabet size $k$, but the
exponent of $2\Lmax+1$ is very much a loose worst-case result.  It applies only
in extreme cases; e.g., when every string spawns its own state, almost all of
which are transient.  In practice, the run time is much shorter than this bound
would lead one to fear.  Average-case results would replace the number of
strings, here bounded by $s(k,\Lmax) \approx k^\Lmax$, with the typical number
of distinct sequences of length $\Lmax$ the process generates with positive
probability, which is $2^{\Lmax\hmu}$ \citep{Cover-and-Thomas}. Similarly,
bounding the number of states by $s(k,\Lmax)$ is generally excessive.

\section{Reliability and Rates of Convergence}

We wish to show that the estimated $\epsilon$-machines produced by the CSSR
algorithm converge to the original process's true $\epsilon$-machine.
Specifically, we will show that the set $\EstCausalStateSet$ of states it
estimates converges in probability to the correct set $\CausalStateSet$ of
causal states: i.e., the probability that
$\EstCausalStateSet \neq \CausalStateSet$
goes to zero as $N \rightarrow \infty$.  To do this, we first look at the
convergence of the empirical conditional frequencies to the true morphs and
then use this to show that the set $\EstCausalStateSet$ of states converges.
Finally, we examine the convergence of predictions once the causal states are
established and say a few words about how, from a dynamical-systems viewpoint,
the $\epsilon$-machine acts as a kind of attractor for the algorithm.

Throughout, we make the following assumptions:
\begin{enumerate}
\item The process is conditionally stationary.  Hence the ergodic theorem for
stationary processes applies \citep[Sec.~9.5]{Grimmett-Stirzaker} and time
averages converge almost surely to the state-space average for whatever ergodic
component the process happens to be in.
\item The original process has only a finite number of causal states.
\item Every state contains at least one suffix of finite length.  That is,
there is some finite $L$ such that every state contains a suffix of length no
more than $L$.  This does not mean that $L$ symbols of history always suffice
to determine the state, just that it is \emph{possible} to synchronize --- in
the sense of \citet{Upper-thesis} --- to every state after seeing no more than
$L$ symbols.
\end{enumerate}

We also assume that CSSR estimates conditional probabilities (morphs) by simple
maximum likelihood.  One can substitute other estimators --- e.g., maximum
entropy --- and only the details of the convergence arguments would change.

\subsection{Convergence of the Empirical Conditional Probabilities to the
Morphs}
\label{sec:convergence-analysis}

The past $\Past$ and the future $\Future$ of the process are independent given
the causal state $\causalstate \in \CausalState$
(Prop. \ref{prop:cond-ind-of-past-and-future}).  More particularly, the past
$\Past$ and the next future symbol $\NextObservable$ are independent given the
causal state. Hence, $\Past$ and $\NextObservable$ are independent, given a
history suffix sufficient to fix us in a cell of the causal-state partition.
Now, the morph for that suffix is a multinomial distribution over $\cal A$,
which gets sampled independently each time the suffix occurs.
Our estimate of the morph is the empirical distribution obtained by IID samples
from that multinomial.  Write this estimated distribution as
$\widehat{\Prob}_N(\NextObservable=a|\PastL = \pastL)$, recalling that
$\Prob(\NextObservable=a|\PastL = \pastL)$ denotes the true morph.  Does
$\widehat{\Prob}_N \rightarrow \Prob$ as $N$ grows?

Since we use the KS test, it is reasonable to employ the variational
distance.\footnote{This metric is compatible with most other standard tests,
too.} Given two distributions $P$ and $Q$ over $\ProcessAlphabet$, the
variational distance is
\begin{equation}
d(P,Q) = \sum_{a\in\ProcessAlphabet}{|P(X=a)-Q(X=a)|} ~.
\end{equation}
Scheff\'{e} showed that
\begin{equation}
d(P,Q) = 2 \max_{\mathrm{A}\in
  2^{\ProcessAlphabet}}{|P(X\in\mathrm{A})-Q(X\in\mathrm{A})|} ~,
\end{equation}
where $2^{\ProcessAlphabet}$ is the power-set of $\ProcessAlphabet$, of
cardinality $2^k$ \citep{Devroye-Lugosi-combinatorial}.

Chernoff's inequality \citep{Vidyasagar-learning} tells us that, if $X_1, X_2,
\ldots X_n$ are IID Bernoulli random variables and $A_n$ is the mean of the
first $n$ of the $X_i$, with probability $\mu$ of success, then
\begin{equation}
{\Prob(|A_n - \mu| \geq t)} \leq 2e^{-2nt^2} ~.
\end{equation}
Applying this to our estimation problem and letting $n = \nu(\pastL)$ be the
number of times we have seen the suffix $\pastL$, we have:
\begin{eqnarray}
\lefteqn{\Prob(d(\widehat{\Prob}_n(\NextObservable|\PastL = \pastL),
\Prob(\NextObservable|\PastL = \pastL)) \geq t) } & & \\
  & = & \Prob\left(t < \sum_{a\in\ProcessAlphabet}
	{\left|\widehat{\Prob}_n(\NextObservable=a|\PastL = \pastL)
	- \Prob(\NextObservable=a|\PastL = \pastL)\right|}\right)\\
  & = & \Prob\left(2t < \max_{\mathrm{A}\in 2^\ProcessAlphabet}
	{\left|\widehat{\Prob}_n(\NextObservable\in\mathrm{A}|\PastL = \pastL)
	- \Prob(\NextObservable\in\mathrm{A}|\PastL = \pastL)\right|}\right) \\
  & \leq & \sum_{\mathrm{A}\in 2^\ProcessAlphabet}
	{\Prob\left(\left|\widehat{\Prob}_n(\NextObservable\in\mathrm{A}|\PastL = \pastL)
	- \Prob(\NextObservable\in\mathrm{A}|\PastL =  \pastL)\right| > 2t\right)} \\
  & \leq & \sum_{\mathrm{A}\in 2^\ProcessAlphabet}{2{e}^{-2n{(2t)}^{2}}}\\
     \label{eqn:first-error-bound}
  & = & 2^{k+1}{e}^{-8n{t}^{2}} ~.
\label{chernoffish-bound-on-history-distribution}
\end{eqnarray}
Thus, the empirical conditional distribution (morph) for each suffix converges
exponentially fast to its true value, as the suffix count $n$ grows.

Suppose there are $s$ suffixes across all the states.  We have seen the $i^{\rm
th}$ suffix $n_i$ times; abbreviate the $s$-tuple $(n_1, \ldots n_s)$ by
$\mathbf{n}$. We want the empirical distribution associated with each suffix to
be close to the empirical distribution of all the other suffixes in its state.
We can ensure this by making all the empirical distributions close to the
state's morph.  In particular, if all distributions are within $t$ of the
morph, then (by the triangle inequality) every distribution is within $2t$ of
every other distribution.  Call the probability that this is not true $q(t,
\mathbf{n})$.  This is the probability that \textit{at least one} of the
empirical distributions is not within $t$ of the state's true morph. $q(t,
\mathbf{n})$ is at most the sum of the probabilities of each suffix being an
outlier.  Hence, if there are $s$ suffixes in total, across all the states,
then the probability that one or more suffixes differs from its true morph by
$t$ or more is
\begin{eqnarray}
q(t, \mathbf{n}) & \leq & \sum_{i=1}^{s}{2k{e}^{-8n_i t^2}}\\
& \leq & 2^{k+1}s{e}^{-8mt^2} ~,
\label{eqn:second-error-bound}
\end{eqnarray}
where $m$ is the least of the $n_i$.  Now, this formula is valid whether we
interpret $s$ as the number of histories actually seen or as the number of
histories needed to infer the true states.  In the first case,
Eq.~(\ref{eqn:second-error-bound}) tells us how accurate we are on the
histories seen; in the latter, over all the ones we need to see. This last is
clearly what we need to prove overall convergence, but knowing $s$, in that
sense, implies more knowledge of the process's structure than we start with.
However, we can upper bound $s$ by the maximum number of morphs possible
--- $s \leq (k^{L+1}-1)/(k-1)$ --- so we only need to know (or, in practice,
pick) $L$.

Which string is least-often seen --- which $n_i$ is $m$ --- generally changes
as more data accumulates.  However, we know that the frequencies of strings
converge almost surely to probabilities as $N \rightarrow \infty$. Since there
are only a finite number of strings of length $L$ or less, it follows that the
empirical frequencies of all strings also converge almost
surely.\footnote{Write $F_i$ for the event that the frequencies of string $i$
\textit{fail} to converge to the probability.  From the ergodic theorem,
$\Prob(F_i) = 0$ for each $i$.  If there are $s$ strings, the event that
represents one or more failures of convergence is $\bigcup_{i=1}^{s}{F_i}$.
Using Bonferroni's inequality, $\Prob(\bigcup_{i=1}^{s}{F_i}) \leq
\sum_{i=1}^{s}{\Prob(F_i)} = s \cdot 0 = 0$.  Hence, all $s$ strings converge
together with probability $1$.} If $p^{*}$ is the probability of the most
improbable string, then $m \rightarrow Np^{*}$ almost surely.  Hence, for
sufficiently large $N$,
\begin{eqnarray}
\label{eqn:third-error-bound}
q(t, \mathbf{n}) & \leq & 2^{k+1}\frac{k^{L}-1}{k-1} e^{-8Np^{*}t^2} ~,
\end{eqnarray}
which vanishes with $N$.

We can go further by invoking the assumption that there are only a finite
number of causal states.  This means that causal states form a finite
irreducible Markov chain.  The empirical distributions over finite sequences
generated by such chains converge exponentially fast to the true distribution
\citep{den-Hollander-large-deviations}. (The rate of convergence depends on
the entropy ${\cal D} (\widehat{\Prob}|\Prob)$ of the empirical distribution
relative to the true.)  Our observed process is a random function of the
process's causal states.  Now, for any pair of distributions $P(X)$ and $Q(X)$
and any measurable function $f: X \rightarrow Y$,
\begin{equation}
d(P(X),Q(X)) \geq d(P(Y),Q(Y)) ~,
\end{equation}
whether $f$ is either a deterministic or a random function. Hence, the
empirical sequence distribution, at any finite length, converges to the true
distribution at least as quickly as the state distributions.  That is, they
converge with at least as large an exponent.  Therefore, under our assumptions,
for each $i$, $n_i/N \rightarrow p_i$ exponentially fast --- there is an
increasing function $r(\varepsilon)$, $r(0) = 0$, and a constant $C$ such that,
for each $\varepsilon \geq 0$,
\begin{eqnarray}
\Prob\left(\left|\frac{n_i}{N} - p_i\right| > \varepsilon\right) & \leq & C e^{-Nr(\varepsilon)} ~.
\end{eqnarray}
($r(\varepsilon)$ is known as the \emph{large-deviation rate function.})  The
probability that $m \leq N(p^{*} - \varepsilon)$ is clearly no more than the
probability that, for one or more $i$, $n_i \leq N(p^{*} - \varepsilon)$.
Since the lowest-probability strings need to make the smallest deviations from
their expectations to cause this, the probability that at least one $n_i$ is
below $N(p^{*} - \varepsilon)$ is no more than $k^L$ times the probability that
the count of the least probable string is below $N(p^{*} - \varepsilon)$.  The
probability that an empirical count $n_i$ is below its expectation by
$\varepsilon$, in turn, is less than the probability that it deviates from its
expectation by $\varepsilon$ in either direction. That is,
\begin{eqnarray}
\Prob(m \leq N(p^{*} - \varepsilon)) & \leq & \Prob\left(\bigcup_{i}{n_i \leq N(p^{*} - \varepsilon)}\right)\\
  & \leq & \sum_{i}{\Prob(n_i \leq N(p^{*} - \varepsilon))}\\
  & \leq & k^L \Prob(n^{*} \leq {N}(p^{*} - \varepsilon))\\
  & \leq & k^L \Prob\left(\left|\frac{n^{*}}{N} - p^{*}\right| > \varepsilon\right)\\
  & \leq & C k^L e^{-Nr(\varepsilon)} ~.
\end{eqnarray}
With a bound on $m$, we can fix an overall exponential bound on the error
probability $q$:
\begin{equation}
q(t,\mathbf{n}) \leq 
  \inf_{\varepsilon \geq 0}{C2^{k+1}k^{L}\frac{k^{L+1} -1}{k-1}
  e^{-8N(p^{*} - \varepsilon) t^2}
  e^{-Nr(\varepsilon)}} ~.
\end{equation}
Solving out the infimum would require knowledge of $r(\epsilon)$, which is
difficult to come by. Whatever $r(\epsilon)$ may be, however, the bound is
still exponential in $N$.

The above bound is crude for small $N$ and especially for small $t$.  In the
limit $t \rightarrow 0$, it tells us that a probability is less than some
positive integer, which, while true, is not at all sharp.  It becomes less
crude, however, as $N$ and $t$ grow.  In any case, it suffices for the present
purposes.

We can estimate $p^{*}$ from the reconstructed $\epsilon$-machine by
calculating its fluctuation spectrum \citep{fluctuation-spectroscopy}.  Young
and Crutchfield demonstrated that the estimates of $p^{*}$ obtained in this way
become quite accurate with remarkably little data, just as estimates of the
entropy rate $\hmu$ do.  At the least, calculating $p^{*}$ provides a
self-consistency check on the reconstructed $\epsilon$-machine.

\subsection{Analysis of Error Probabilities}

Let us consider the kinds of statistical error each of the algorithm's
three procedures can produce.

Since it merely sets up parameters and data structures, nothing goes wrong
in Procedure I (Initialize).

Procedure II (Homogenize) can make two kinds of errors.  First, it can group
together histories with different distributions for the next symbol.  Second,
it can fail to group together histories that have the same distribution.  We
will analyze both cases together.

It is convenient here to introduce an additional term.  By analogy with the
causal states, the \emph{precausal states} are defined by the following
equivalence relation: Two histories are precausally equivalent when they have
the same morph.  The precausal states are then the coarsest states
(partition) that are weakly prescient.  The causal states are either the same
as the precausal states or a refinement of them.  Procedure II ought to deliver
the correct partition of histories into precausal states.

Suppose $\past_i$ and $\past_j$ are suffixes in the same state, with counts
$n_i$ and $n_j$.  No matter how large their counts, there is always
some variational distance $t$ such that the significance test will not separate
estimated distributions differing by $t$ or less.  If we make $n_i$ large
enough, then, with probability arbitrarily close to one, the estimated
distribution for $\past_i$ is within $t/2$ of the true morph, and similarly for
$\past_j$.  Thus, the estimated morphs for the two suffixes are within $t$ of
each other and will be merged.  Indeed, if a state contains any finite number
of suffixes, by obtaining a sufficiently large sample of each, we can ensure
(with arbitrarily high probability) that they are all within $t/2$ of the true
morph and so within $t$ of each other and thus merged.  In this way, the
probability of inappropriate splitting can be made arbitrarily small.

If each suffix's conditional distribution is sufficiently close to its true
morph, then any well behaved test will eventually separate suffixes that belong
to different morphs.  More concretely, let the variational distance between the
morphs for the pair of states $\sigma_i$ and $\sigma_j$ be $d_{ij}$.  A well
behaved test will distinguish two samples whose distance is some constant
fraction of this or more --- say, $3d_{ij}/4$ or more --- if $n_i$ and $n_j$
are large enough.  By taking $n_i$ large enough, we can make sure, with
probability arbitrarily close to $1$, that the estimated distribution for
$\sigma_i$ is within some small distance of its true value --- say, $d_{ij}/8$.
We can do the same for $\sigma_j$ by taking $n_j$ large enough.  Therefore,
with probability arbitrarily close to one, the distance between the estimated
morphs for $\sigma_i$ and $\sigma_j$ is at least $3d_{ij}/4$, and $i$ and $j$
are, appropriately, separated.  Hence, the probability of erroneous
non-separations can be made as small as desired.

Therefore, by taking $N$ large enough, we can make the probability that the
correct precausal states are inferred arbitrarily close to $1$.

If every history is correctly assigned to a precausal state, then nothing can
go wrong in Procedure III (Determinize).  Take any pair of histories $\past_1$
and $\past_2$ in the same precausal state: either they belong to the same
causal state or they do not.  If they do belong to the same causal state, then
by determinism, for every string $w$, $\past_1 w$ and $\past_2 w$ belong to the
same causal state.  Since the causal states are refinements of the precausal
states, this means that $\past_1 w$ and $\past_2 w$ also belong to the same
precausal state.  Contrarily, if $\past_1$ and $\past_2$ belong to different
causal states, they must give different probabilities for \emph{some} strings.
Pick the shortest such string $w$ (or any one of them, if there is more than
one) and write it as $w = va$, where $a$ is a single symbol.  Then the
probability of $a$ depends on whether we saw $\past_1 v$ or $\past_2
v$.\footnote{Otherwise,$\past_1$ and $\past_2$ must assign different
  probabilities to $v$, and so $w = va$ is not the shortest string on which
  they differ.}  So $\past_1 v$ and $\past_2 v$ have distinct morphs and belong
to different precausal states.  Hence, determinization will separate $\past_1$
and $\past_2$, since, by hypothesis, the precausal states are correct.  Thus,
histories will always be separated, if they should be, and never, if they
should not.

Since determinization always refines the partition with which it starts and the
causal states are a refinement of the precausal states, there is no chance of
merging histories that do not belong together.  Hence, Procedure III will
always deliver the causal states, if it starts with the precausal states.  We
will not examine the question of whether Procedure III can rectify a mistake
made in Procedure II.  Experientially, this depends on the precise way
determinization is carried out and, most typically, if the estimate of the
precausal states is seriously wrong, determinization only compounds the
mistakes.  Procedure III does not, however, enhance the \textit{probability} of
error.

To conclude: If the number of causal states is finite and $\Lmax$ is
sufficiently large, the probability that the states estimated are not the
causal states becomes arbitrarily small, for sufficiently large $N$.  Hence the
CSSR algorithm, considered as an estimator, is (i) consistent \citep{Cramer},
(ii) probably approximately correct \citep{Kearns-Vazirani,Vapnik-nature}, and
(iii) reliable \citep{Spirtes-Glymour-Scheines,Kelly-reliable}, depending on
which field one comes from.

\subsection{Dynamics of the Learning Algorithm}
\label{sec:dyn-learn}

\begin{figure}
\begin{center}
\resizebox{!}{3in}{\includegraphics{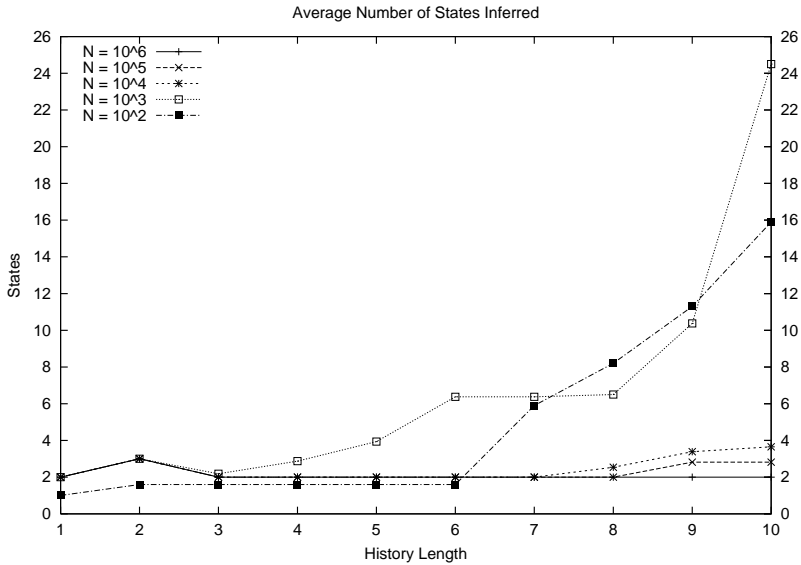}}
\end{center}
\caption{Number of states inferred as a function of history length $\Lmax$ and
  data size $N$ for the even process of Section \ref{sec:even-process-example}.
  The true number of states is 2.  Curves are averages over 30 independent
  realizations at each $N$.  Here, and in subsequent figures, the significance
  level $\alpha$ was fixed to ${10}^{-3}$.}
\label{figure:EP-number-states}
\end{figure}

We may consider the CSSR algorithm itself to be a stochastic dynamical system,
moving through a state space of possible history-partitions or
$\epsilon$-machines. What we have seen is that the probability that CSSR does
not infer the correct states (partition) --- that it does not have the causal
architecture right --- drops exponentially as time goes on.  By the
Borel-Cantelli lemma, therefore, CSSR outputs $\epsilon$-machines that have the
wrong architecture only finitely many times before fixing on the correct
architecture forever. Thus, the proper architecture is a kind of absorbing
region in $\epsilon$-machine space.  In fact, almost all of the algorithm's
trajectories end up there and stay. (There may be other absorbing regions, but
only a measure-$0$ set of inference trajectories reach them.)

Because CSSR can only create new states as it steps through $L$ to $\Lmax$, it
is tempting to conjecture that CSSR ``converges from below'' --- that the
number of states grows monotonically from one to the true value, perhaps for
growing $N$ with fixed $\Lmax$, or for growing $\Lmax$ with fixed $N$.  This is
not so, however.  If $\Lmax$ is too large relative to $N$ (see Section
\ref{sec:Parameters}), the probabilities of long strings will not be
consistently estimated, so the hypothesis tests in Procedure II become
unreliable and tend to produce too many states.  Procedure III then amplifies
this excess.  More broadly, after Procedure II makes the precausal states,
Procedure III determinizes them, and the deterministic version of an incorrect
set of states (e.g., from setting $\Lmax$ too small) can easily be larger than
the correct $\epsilon$-machine.  The general situation, illustrated by Figure
\ref{figure:EP-number-states}, does not allow us to make any generalization
about convergence ``from above'' or ``from below''.

\begin{figure}[p]
\begin{center}
\resizebox{!}{2.92in}{\includegraphics{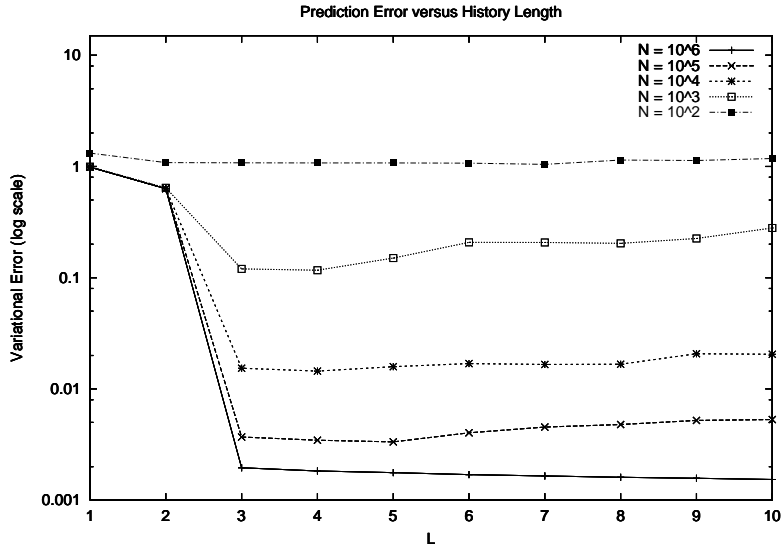}}
\end{center}
\caption{Prediction error (logarithmic scale) as a function of history length
  and data size.  The figure depicts the total-variation distance between the
  actual distribution over words of length 10, and that implied by the inferred
  model (true generalization error).  Error is shown as a function of
  $\Lmax$ (running from 1 to 10) and $N$ (running from ${10}^{2}$ to
  ${10}^{6}$), averaged over 30 independent realizations at each $N$.}
\label{figure:EP-error}
\end{figure}

\begin{figure}[p]
\begin{center}
\resizebox{!}{2.92in}{\includegraphics{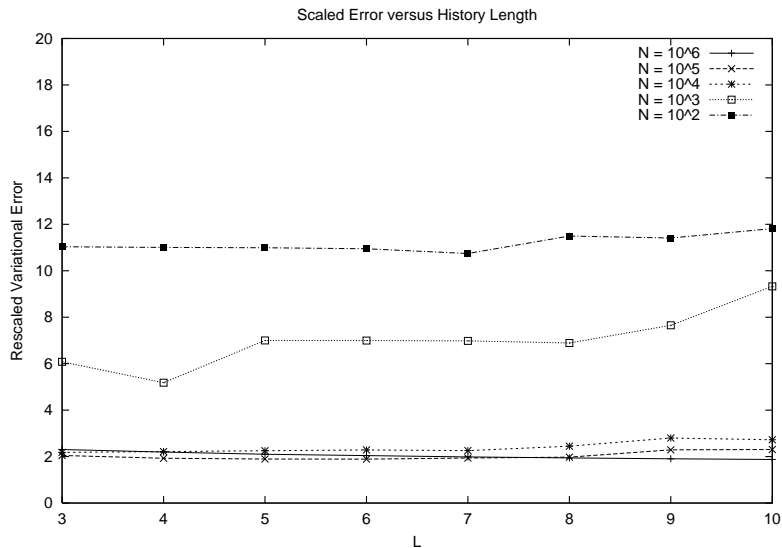}}
\end{center}
\caption{Scaling of prediction error (linear scale) as a function of $N$.  As
  in Figure \ref{figure:EP-error}, but the variational distance is now scaled
  by a factor of $\sqrt{N}$.  For clarity, we have restricted $\Lmax$ to the
  range 3--10, since 3 is the smallest value of $\Lmax$ for which CSSR can find
  the correct model.  At this significance level, CSSR never finds
  the correct model for $N = {10}^{2}$, and only sporadically for $N =
  {10}^3$, hence those lines do not fall on the scaling curve.}
\label{figure:EP-error-scaling}
\end{figure}

We are not satisfied with getting the correct causal architecture, however. We
also want the morphs to converge. Here we can exploit the fact that the
estimated morph for each causal state is an average over the histories it
contains. If there are $s_c$ causal states and the least frequently sampled one
has been seen $m_c$ times, then reasoning parallel to that in Section
\ref{sec:convergence-analysis} above tells us that the probability any of our
estimated morphs differs from its true morph by $t$ or more is at most
$2^{k+1}s_c exp(-8m_c t^2)$.  Moreover, since the causal states form an
irreducible Markov chain, $m_c$ will converge exponentially quickly to
$Np^{*}_c$.

As a general note, while the probability of errors of a given size goes down
exponentially with $N$, this does not imply that the expectation of the error
is exponentially decreasing.  Rather, the expected variational difference
between the true and empirical distributions for a multinomial goes as
$N^{-1/2}$ \citep{Devroye-Lugosi-combinatorial}.  Since $N$ appears in the form
for the error probability only in the combination $Np^{*}_c$, we expect a
variational error scaling as $1/\sqrt{Np^{*}_c}$.  Readers may judge the
quality of the data-collapse of the rescaled error for themselves from Figures
\ref{figure:EP-error} and \ref{figure:EP-error-scaling}.

\subsection{Choice of Algorithm Parameters}
\label{sec:Parameters}

CSSR has two adjustable parameters: the maximum history length used, $\Lmax$,
and the significance level $\alpha$.

We have seen that there is a lower bound for the acceptable value of $\Lmax$,
namely, it must be large enough that every state contains at least one suffix
of that length.  Let us call this least acceptable value $\Lmin$.  If $\Lmax <
\Lmin$, our proof of convergence fails, and in general neither CSSR nor any
other procedure can reconstruct the causal states using histories of length
$\Lmax$.  For periodic processes, for instance, $\Lmin$ is equal to the period.
Since the statistical complexity $\Cmu$ of a process with period $p$ is just
$\log{p}$, this suggests that generally $\Lmin \approx 2^\Cmu$, but this is
unproven, and we know of no estimator for $\Cmu$ other than direct
reconstruction.

Rather than try to determine $\Lmin$, one might use the largest $\Lmax$
compatible with available memory (and one's own impatience).  However, the size
of the data-set $N$ limits the permissible values of $\Lmax$.  We can see this
as follows.  Estimating the distribution for the next symbol conditional on a
history of length $L-1$ is equivalent to estimating the distribution of strings
of length $L$.  Now, the asymptotic equipartition property
\citep{Cover-and-Thomas} tells us that the number of strings which appear with
positive probability grows as $2^{L\hmu}$.  That is, the number of
probabilities we need to estimate grows exponentially with $L$.  To get an
adequate number of samples per string thus requires an exponentially growing
amount of data.  Conversely, adequate sampling limits $L$ to be, at most, on
the order of the logarithm of $N$.

A result of \citet{Marton-Shields} makes this more precise.  Let $L(N)$ be the
the maximum $L$ we can use when we have $N$ data-points.  If the observed
process satisfies the weak Bernoulli property (which random functions of
irreducible Markov chains do), then a sufficient condition for the convergence
of sequence probability estimates is that $L(N) \leq \log{N}/(\hmu +
\varepsilon)$, for some positive $\varepsilon$.  One might wonder if cleverness
could improve this; but Marton and Shields also showed that, if $L(N) \geq
\log{N}/\hmu$, probability estimates over length $L$ words do not converge.  We
must know $\hmu$ to use this result, but $\log{k} \geq \hmu$, so we can err on
the side of caution by substituting the log of the alphabet size for the
entropy rate.  For a given process and data-set, it is of course possible that
$L(N) < \Lmin$, in which case we simply haven't enough data to reconstruct the
true model.

As to the significance level $\alpha$, the proof of convergence shows us that
it doesn't matter, at least asymptotically in $N$: histories with distinct
morphs will be put in separate states, and histories with the same morph will
be joined together in the same state.  In the meanwhile, with finite data, it
does affect the kind of error CSSR makes.  The significance level is the
probability that two samples, drawn from the same distribution, would be split
apart by the test.  If $\alpha$ is small, therefore, we will split only on
large values of the test statistic --- only (as it were) glaringly obvious
distinctions between empirical distributions will lead to splits.  To become
sensitive to very small distinctions between states, we need either large
quantities of data, or a high significance level, which would make us
more likely to split accidentally.  The significance level therefore
indicates our willingness to risk seeing structure that isn't really there,
i.e., creating states due merely to sampling error.

\section{Comparison with Earlier Techniques}

In this section, we compare CSSR to two classes of existing techniques for
pattern discovery in time series: ``context'' or variable-length Markov models,
and state-merging methods for causal state reconstruction.  We see that the
``context'' methods are actually included as a special case of causal state
reconstruction, under restrictive, and generally under-appreciated, assumptions.
Merging methods for state reconstruction, while they have the same domain of
applicability as CSSR, are less well-behaved, and converge more slowly.

\subsection{Variable-Length Markov Models}
\label{sec:VLMMs}

The ``context'' algorithm of \citet{Rissanen-1983} and its descendants
\citep{Buhlmann-Wyner,Willems-Shtarkov-Tjalkens-CTW,Tino-Dorffner-fractal}
construct so-called ``variable-length Markov models'' (VLMMs) from sequence
data.  The object is to find a set of \emph{contexts} such that, given the
context, the past of the sequence and its next symbol are conditionally
independent.  Contexts are taken to be suffixes of the history, and the
algorithms work by examining increasingly long histories, creating new contexts
by splitting existing ones into longer suffixes when thresholds of error are
exceeded.  (This means that contexts can be arranged in a tree, so these are
also called ``context tree'' or ``probabilistic suffix tree'' algorithms.)

This description of variable-length Markov model algorithms makes it clear that
they are related to causal state reconstruction, and, like reconstruction
algorithms, they infer both architecture and parameters.  However, causal
state methods have several important advantages over VLMM methods.  First, we
use the known properties of the states we are looking for to guide search.
Second, rather than creating states when we cross some arbitrary threshold for,
say, the Kullback-Leibler divergence, we use well-understood statistical tests
for identity of distribution.  Third, context methods suffer from the following
defect.  Each state they infer is represented by a single suffix.  That is,
their states consist of all and only the histories ending in a particular
suffix.  For many processes, the causal states contain multiple suffixes.  In
these cases, multiple ``contexts'' are needed to represent a single causal
state, so VLMMs are generally more complicated than $\epsilon$-machines.

Recall the even process of Section \ref{sec:even-process-example}.  It has
two recurrent causal states, labeled A and B in Figure
\ref{figure:EvenProcess}.  Any history terminated by a 0, or by a 0 followed by
an even number of 1s, belongs to state A.  Any history terminated by a 0
followed by an odd number of 1s, belongs to B.  Clearly A and B both contain
infinitely many suffixes, and so correspond to an infinite number of contexts.
VLMM algorithms are simply incapable of capturing this structure.  If
we let $\Lmax$ grow, a VLMM algorithm will increase the number of
contexts it finds without bound, but cannot achieve the same
combination of predictive power and model simplicity as causal state
reconstruction.

Moreover, the even process is not an isolated pathological case.  Rather, it is
a prototype of the \emph{strictly sofic processes}
\citep{Weiss-1973,Badii-Politi}, which are, in essence, processes which can be
described as hidden Markov models, but are not Markov chains of any finite
order\footnote{More exactly, for each history $\past$, the \emph{follower set}
  consists of the futures $\future$ which can succeed it.  A process is
  \emph{sofic} if it has only a finite number of different follower sets, and
  strictly sofic if it is sofic and has an infinite number of irreducible
  forbidden words.}.  VLMMs cannot capture strictly sofic processes,
for the same reason they cannot capture the even process.  However, sofic
processes are simply regular languages, since they have only finitely many
states.  Causal states cannot provide a finite representation of every regular
language \citep{Upper-thesis,Calculi-of-emergence}, but the class they capture
strictly includes those captured by VLMMs.

\subsection{State-Merging $\epsilon$-Machine Inference Algorithms}
\label{sec:tree-merging-algorithm}

Existing $\epsilon$-machine reconstruction procedures use what one might
call state compression or merging.  The default assumption is that each
distinct history encountered in the data is a distinct causal state.  Histories
are then merged into causal states when their morphs are close.  Kindred
merging procedures can learn hidden Markov models \citep{Stolcke-Omohundro} and
finite automata \citep{Trakhtenbrot-and-Barzdin,Murphy-passively-learning}.

The standard $\epsilon$-machine inference algorithm is the subtree-merging
algorithm introduced by Crutchfield and Young
\citep{Inferring-stat-compl,Computation-at-the-Onset}.  The algorithm begins by
building a $k$-ary tree of some pre-set depth $D$, where paths through the tree
correspond to sequences of observations of length $D$, obtained by sliding a
length-$D$ window through the data stream (or streams, if there are several).
If $D = 4$, say, and the sequence $abba$ is encountered, the path in the tree
will start at the root node, take the edge labeled $a$ to a new node, then take
the outgoing edge labeled $b$ to a third node, then the edge labeled $b$ from
that, and finally the edge labeled $a$ to a fifth node, which is a leaf.  An
edge of the tree is labeled, not just with a symbol, but also with the number
of times that edge has been traversed in scanning through the data stream.
Denote by $\nu(a_i|n)$ the number on the $a_i$ edge going out of node $n$ and
$N$ the total number of sequences entered into the tree.  The tree is a
$D$-block Markov model of the process.  Each level $l$ gives an estimate of the
distribution of length-$l$ words in the data stream.

The traversal-counts are converted into empirical conditional probabilities by
normalization:
\begin{eqnarray*}
\ProbEst(a_i|n) & = & \frac{\nu(a_i|n)}{\sum_{a_j}{\nu(a_j|n)}} ~.
\end{eqnarray*}
Thus, attached to each non-leaf node is an empirical conditional distribution
for the next symbol.  If node $n$ has descendants to depth $K$, then it has (by
implication) a conditional distribution for futures of length $K$.

The merging procedure is now as follows.  Consider all nodes with subtrees of
depth $L = D/2$.  Take any two of them.  If all the
empirical probabilities attached to the length-$L$ path in their subtrees are
within some constant $\delta$ of one another, then the two nodes are
equivalent.  They should be merged with one another.  The new node for the root
will have incoming links from both the parents of the old nodes.  This
procedure is to be repeated until no further merging is
possible.\footnote{Since the criterion for merging is not a true equivalence
relation (it lacks transitivity), the order in which states are examined for
merging matters, and various tricks exist for dealing with this.  See, e.g.,
\citet{Hanson-thesis}.}  It is clear that, given enough data, a long enough
$L$, and a small enough $\delta$, the subtree algorithm will converge on the
causal states.

All other methods for $\epsilon$-machine reconstruction currently in use are
also based on merging.  Take, for instance, the ``topological'' or ``modal''
merging procedure of \citet{Perry-Binder-finite-stat-compl}.  They consider the
relationship between histories and futures, both (in the implementation) of
length $L$.  Two histories are assigned to the same state if the sets of
futures that can succeed them are identical.\footnote{This \textit{is} an
equivalence relation, but it is not causal equivalence.}  The distribution over
those futures is then estimated for each state, not for each history.

As we said, the default assumption of current state-merging methods is that
each history is its own state.  The implicit null model of the process is thus
the most complex possible, given the length of histories available, and is
whittled down by merging.   In all this, they are the opposite of CSSR. State
splitting starts by putting every history in one state --- a zero-complexity
null model that is elaborated by splitting.

Unfortunately, state-merging has inherent difficulties.  For instance: what is
a reasonable value of morph similarity $\delta$? Clearly, as the amount of data
increases and the law of large numbers makes empirical probabilities converge
to true probabilities, $\delta$ should grow smaller.  But it is grossly
impractical to calculate what $\delta$ should be, since the null model itself
is so complicated.  (Current best practice is to pick $\delta$ as though the
process were an IID multinomial, which is the opposite of the algorithm's
implicit null model.)  In fact, using the same $\delta$ for every pair of tree
nodes is unreliable.  The nodes will not have been visited equally often, being
associated with different tree depths, so the conditional probabilities in
their subtrees vary in accuracy.  An analysis of the convergence of empirical
distributions, of the kind we made in Section \ref{sec:convergence-analysis}
above, could give us a handle on $\delta$, but reveals another difficulty.
CSSR must estimate $2^k$ probabilities for each history --- one for each member
of the power-set of the alphabet.  The subtree-merging algorithm, however, must
estimate the probability of each member of the power set of future sequences,
i.e., $2^{k^L}$ probabilities.  This is an exponentially larger number, and the
corresponding error bounds would be worse by this factor.

The theorems in \citet{CMPPSS} say a great deal about the causal states: they
are deterministic, they are Markovian, and so on.  No previous reconstruction
algorithm made use of this information to guide its search. Subtree-merging
algorithms can return nondeterministic states, for instance, which cannot
possibly be the true causal states.\footnote{It is sometimes claimed
\citep{Palmer-inference-versus-imprint} that the nondeterminism is due to
nonstationarity in the data stream.  While a nonstationary source can cause the
subtree-merging algorithm to return nondeterministic states, the algorithm is
quite capable of doing this when the source is IID.}  While the subtree
algorithm converges, and other merging algorithms probably do too, CSSR should
do better, both in terms of the kind of result it delivers and the rate at
which it approaches the correct result.

\section{Conclusion}

We briefly map out directions for future exploration, and finish by summarizing
our results.

\subsection{Future Directions}

A number of directions for future work present themselves.  Elsewhere, we have
developed extensions of computational mechanics to transducers and interacting
time series and to spatio-temporal dynamical systems
\citep{Comp-mech-of-CA-example,CRS-thesis,Bottleneck-note}.
It is clear that the present algorithm can be applied to transducers, and we
feel that it can be applied to spatio-temporal systems.

One can easily extend the \emph{formalism} of computational mechanics to
processes that take on continuous values at discrete times, but this has never
been implemented.  Much of the machinery we employed here carries over
straightforwardly to the continuous setting, e.g., empirical process theory for
IID samples \citep{Devroye-Lugosi-combinatorial} or the Kolmogorov-Smirnov
test.  The main obstacle to simply using CSSR in its present form is the need
for continuous interpolation between the (necessarily finite) measurements.
However, all methods of predicting continuous-valued processes must likewise
impose some interpolation scheme.  It seems likely that schemes along the lines
of those used by \citet{Bosq-nonparametric} or
\citet{Fraser-Dimitriadis-forecasting-densities} would work with CSSR.
Continuous-valued, continuous-time processes raise more difficult questions,
which we shall not even attempt to sketch here.

CSSR currently returns a single model, and so provides a ``point estimate'' of
the causal states.  This raises the question of what the corresponding
confidence region would look like, and how it might be computed.  Similarly, an
estimate of the expected degree of over-fitting, as a function either of the
number of states or of $\Cmu$, would open the way to applying the structural
risk minimization principle \citep{Vapnik-nature}.

Potentially, $\epsilon$-machines and our algorithm can be applied in any domain
where HMMs have proved their value (e.g., bioinformatics
\citep{Baldi-Brunak-bioinfo}) or where there are poorly-understood processes
generating sequential data, such as speech, in which one wishes to find
non-obvious or very complex patterns.

\subsection{Summary}

We have presented a new algorithm for pattern discovery in time series.  Given
samples of a conditionally stationary process, the algorithm reliably infers
the process's causal architecture.  Under certain conditions on the process,
not uncommonly met in practice, the algorithm almost surely returns an
incorrect architecture only finitely many times.  The time complexity of the
algorithm is linear in the data size.  We have proved it works reliably on all
processes with finitely many causal states.  Finally, we have argued that CSSR
will consistently outperform prior causal-state-merging algorithms and
context-tree methods.

\section*{Acknowledgments}

Our work has been supported by the Dynamics of Learning project at SFI, under
DARPA cooperative agreement F30602-00-2-0583, and by the Network Dynamics
Program here, which is supported by Intel Corporation.  KLS received support
during summer 2000 from the NSF Research Experience for Undergraduates Program.
We thank Erik van Nimwegen for providing us with a preprint of
\citet*{Bussemaker-et-al-genome-dictionary}, and suggesting that similar
procedures might be used to infer causal states.  We thank Dave Albers,
P.-M. Binder, Dave Feldman, Rob Haslinger, Cris Moore, Jay Palmer, Eric Smith,
and Dowman Varn for valuable conversation and correspondence, K. Kedi for moral
support in programming and writing, and Ginger Richardson for initiating our
collaboration.

\appendix

\section{Information Theory}

Our notation and terminology follows that of \citet{Cover-and-Thomas}.

Given a random variable $X$ taking values in a discrete set $\mathcal{A}$,
the \textit{entropy $H[X]$ of $X$} is
\begin{eqnarray*}
H[X] & \equiv & -\sum_{a\in\mathcal{A}}{\Prob(X=a)\log_2{\Prob(X=a)}} ~.
\end{eqnarray*}
$H[X]$ is the expectation value of $-\log_2{\Prob(X)}$.  It represents
the uncertainty in $X$, interpreted as the mean number of binary distinctions
(bits) needed to identify the value of $X$.  Alternately, it is the minimum
number of bits needed to encode or describe $X$.  Note that $H[X] = 0$ if
and only if $X$ is (almost surely) constant.

The \textit{joint entropy} $H[X,Y]$ of two variables $X$ and $Y$ is the
entropy of their joint distribution:
\begin{eqnarray*}
H[X,Y] & \equiv & -\sum_{a\in\mathcal{A},b\in\mathcal{B}}{\Prob(X=a,Y=b)\log_2{\Prob(X=a,Y=b)}} ~.
\end{eqnarray*}

The \textit{conditional entropy} of $X$ given $Y$ is
\begin{eqnarray*}
H[X|Y] & \equiv & H[X,Y] - H[Y] ~.
\end{eqnarray*}
$H[X|Y]$ is the average uncertainty remaining in $X$, given a knowledge of $Y$.

The \textit{mutual information} $I[X;Y]$ between $X$ and $Y$ is
\begin{eqnarray*}
I[X;Y] & \equiv &  H[X] - H[X|Y] ~.
\end{eqnarray*}
It gives the reduction in $X$'s uncertainty due to knowledge of $Y$ and
is symmetric in $X$ and $Y$.

The \emph{entropy rate} $\hmu$ of a stochastic process
$\ldots, S_{-2}, S_{-1}, S_0, S_1, S_2, \ldots$ is
\begin{eqnarray*}
\hmu & \equiv & \lim_{\LLimit}{H[\NextObservable|\PastL]}\\
& = & H[\NextObservable|\Past] ~.
\end{eqnarray*}
(The limit always exists for conditionally stationary processes.)  $\hmu$
measures the process's unpredictability, in the sense that it is the
uncertainty which remains in the next measurement even given complete
knowledge of the past.

 \bibliography{locusts}

\begin{thebibliography}{50}
\expandafter\ifx\csname natexlab\endcsname\relax\def\natexlab#1{#1}\fi
\expandafter\ifx\csname url\endcsname\relax
  \def\url#1{{\tt #1}}\fi

\bibitem[Badii and Politi(1997)]{Badii-Politi}
Remo Badii and Antonio Politi.
\newblock {\em Complexity: {Hierarchical} Structures and Scaling in Physics},
  volume~6 of {\em Cambridge Nonlinear Science Series}.
\newblock Cambridge University Press, Cambridge, 1997.

\bibitem[Baldi and Brunak(1998)]{Baldi-Brunak-bioinfo}
Pierre Baldi and S{\o}ren Brunak.
\newblock {\em Bioinformatics: {The} Machine Learning Approach}.
\newblock Adaptive Computation and Machine Learning. MIT Press, Cambridge,
  Massachusetts, 1998.

\bibitem[Blackwell and Girshick(1954)]{Blackwell-Girshick}
David Blackwell and M.~A. Girshick.
\newblock {\em Theory of Games and Statistical Decisions}.
\newblock Wiley, New York, 1954.
\newblock Reprinted New York: Dover Books, 1979.

\bibitem[Blackwell and Koopmans(1957)]{Blackwell-identifiability}
David Blackwell and Lambert Koopmans.
\newblock On the identifiability problem for functions of finite {Markov}
  chains.
\newblock {\em Annals of Mathematical Statistics}, 28:\penalty0 1011--1015,
  1957.

\bibitem[Bosq(1998)]{Bosq-nonparametric}
Denis Bosq.
\newblock {\em Nonparametric Statistics for Stochastic Processes: {Estimation}
  and Prediction}, volume 110 of {\em Lecture Notes in Statistics}.
\newblock Springer-Verlag, Berlin, 2nd edition, 1998.

\bibitem[B{\"u}hlmann and Wyner(1999)]{Buhlmann-Wyner}
Peter B{\"u}hlmann and Abraham~J. Wyner.
\newblock Variable length {Markov} chains.
\newblock {\em Annals of Statistics}, 27:\penalty0 480--513, 1999.
\newblock URL \url{http://www.stat.berkeley.edu/tech-reports/479.abstract1}.

\bibitem[Bussemaker et~al.(2000)Bussemaker, Li, and
  Siggia]{Bussemaker-et-al-genome-dictionary}
Harmen~J. Bussemaker, Hao Li, and Eric~D. Siggia.
\newblock Building a dictionary for genomes: {Identification} of presumptive
  regulatory sites by statistical analysis.
\newblock {\em Proceedings of the National Academy of Sciences USA},
  97:\penalty0 10096--10100, 2000.
\newblock URL \url{http://www.pnas.org/cgi/doi/10/1073/pnas.180265397}.

\bibitem[Clarke et~al.(2001)Clarke, Freeman, and
  Watkins]{Watkins-et-al-comp-mech-of-geomag}
Richard~W. Clarke, Mervyn~P. Freeman, and Nicholas~W. Watkins.
\newblock The application of computational mechanics to the analysis of
  geomagnetic data.
\newblock {\em Physical Review E}, submitted, 2001.
\newblock E-print, arxiv.org, cond-mat/0110228.

\bibitem[Cover and Thomas(1991)]{Cover-and-Thomas}
Thomas~M. Cover and Joy~A. Thomas.
\newblock {\em Elements of Information Theory}.
\newblock Wiley, New York, 1991.

\bibitem[Cram{\'e}r(1945)]{Cramer}
Harald Cram{\'e}r.
\newblock {\em Mathematical Methods of Statistics}.
\newblock Almqvist and Wiksells, Uppsala, 1945.
\newblock Republished by Princeton University Press, 1946, as vol. 9 in the
  Princeton Mathematics Series and as a paperback in the Princeton Landmarks in
  Mathematics and Physics series, 1999.

\bibitem[Crutchfield(1994)]{Calculi-of-emergence}
James~P. Crutchfield.
\newblock The calculi of emergence: {C}omputation, dynamics, and induction.
\newblock {\em Physica D}, 75:\penalty0 11--54, 1994.

\bibitem[Crutchfield and Young(1989)]{Inferring-stat-compl}
James~P. Crutchfield and Karl Young.
\newblock Inferring statistical complexity.
\newblock {\em Physical Review Letters}, 63:\penalty0 105--108, 1989.

\bibitem[Crutchfield and Young(1990)]{Computation-at-the-Onset}
James~P. Crutchfield and Karl Young.
\newblock Computation at the onset of chaos.
\newblock In Wojciech~H. Zurek, editor, {\em Complexity, Entropy, and the
  Physics of Information}, volume~8 of {\em Santa Fe Institute Studies in the
  Sciences of Complexity}, pages 223--269, Reading, Massachusetts, 1990.
  Addison-Wesley.

\bibitem[den Hollander(2000)]{den-Hollander-large-deviations}
Frank den Hollander.
\newblock {\em Large Deviations}, volume~14 of {\em Fields Institute
  Monographs}.
\newblock American Mathematical Society, Providence, Rhode Island, 2000.

\bibitem[Devroye and Lugosi(2001)]{Devroye-Lugosi-combinatorial}
Luc Devroye and G{\'a}bor Lugosi.
\newblock {\em Combinatorial Methods in Density Estimation}.
\newblock Springer Series in Statistics. Springer-Verlag, Berlin, 2001.

\bibitem[Feldman and Crutchfield(1998)]{DNCO}
David~P. Feldman and James~P. Crutchfield.
\newblock Discovering noncritical organization: {S}tatistical mechanical,
  information theoretic, and computational views of patterns in simple
  one-dimensional spin systems.
\newblock {\em Journal of Statistical Physics}, submitted, 1998.
\newblock URL \url{http://www.santafe.edu/projects/CompMech/papers/DNCO.html}.
\newblock Santa Fe Institute Working Paper 98-04-026.

\bibitem[Fraser and
  Dimitriadis(1993)]{Fraser-Dimitriadis-forecasting-densities}
Andrew~M. Fraser and Alexis Dimitriadis.
\newblock Forecasting probability densities by using hidden {Markov} models
  with mixed states.
\newblock In A.~S. Weigend and N.~A. Gershenfeld, editors, {\em Time Series
  Prediction: {Forecasting} the Future and Understanding the Past}, volume~15
  of {\em SFI Studies in the Sciences of Complexity}, pages 265--282, Reading,
  Massachusetts, 1993. Addison-Wesley.

\bibitem[Grimmett and Stirzaker(1992)]{Grimmett-Stirzaker}
G.~R. Grimmett and D.~R. Stirzaker.
\newblock {\em Probability and Random Processes}.
\newblock Oxford University Press, Oxford, 2nd edition, 1992.

\bibitem[Hand et~al.(2001)Hand, Mannila, and Smyth]{Hand-Mannila-Smyth}
David Hand, Heikki Mannila, and Padhraic Smyth.
\newblock {\em Principles of Data Mining}.
\newblock Adaptive Computation and Machine Learning. MIT Press, Cambridge,
  Massachusetts, 2001.

\bibitem[Hanson(1993)]{Hanson-thesis}
James~E. Hanson.
\newblock {\em Computational Mechanics of Cellular Automata}.
\newblock PhD thesis, University of California, Berkeley, 1993.
\newblock URL \url{http://www.santafe.edu/projects/CompMech/}.

\bibitem[Hanson and Crutchfield(1997)]{Comp-mech-of-CA-example}
James~E. Hanson and James~P. Crutchfield.
\newblock Computational mechanics of cellular automata: {A}n example.
\newblock {\em Physica D}, 103:\penalty0 169--189, 1997.

\bibitem[Hastie et~al.(2001)Hastie, Tibshirani, and
  Friedman]{Hastie-Tibshirani-Friedman}
Trevor Hastie, Robert Tibshirani, and Jerome Friedman.
\newblock {\em The Elements of Statistical Learning: Data Mining, Inference,
  and Prediction}.
\newblock Springer Series in Statistics. Springer, New York, 2001.

\bibitem[Hinton and Sejnowski(1999)]{Neural-Computation-unsupervised}
Geoffrey Hinton and Terrence~J. Sejnowski, editors.
\newblock {\em Unsupervised Learning: {Foundations} of Neural Computation}.
\newblock Computational Neuroscience. MIT Press, Cambridge, Massachusetts,
  1999.

\bibitem[Hollander and Wolfe(1999)]{Hollander-Wolfe-nonparametric-methods}
Myles Hollander and Douglas~A. Wolfe.
\newblock {\em Nonparametric Statistical Methods}.
\newblock Wiley Series in Probability and Statistics: Applied Probability and
  Statistic. Wiley, New York, 2nd edition, 1999.

\bibitem[Kearns and Vazirani(1994)]{Kearns-Vazirani}
Michael~J. Kearns and Umesh~V. Vazirani.
\newblock {\em An Introduction to Computational Learning Theory}.
\newblock MIT Press, Cambridge, Massachusetts, 1994.

\bibitem[Kelly(1996)]{Kelly-reliable}
Kevin~T. Kelly.
\newblock {\em The Logic of Reliable Inquiry}, volume~2 of {\em Logic and
  Computation in Philosophy}.
\newblock Oxford University Press, Oxford, 1996.

\bibitem[Lind and Marcus(1995)]{Lind-Marcus}
Douglas Lind and Brian Marcus.
\newblock {\em An Introduction to Symbolic Dynamics and Coding}.
\newblock Cambridge University Press, Cambridge, England, 1995.

\bibitem[Marton and Shields(1994)]{Marton-Shields}
Katalin Marton and Paul~C. Shields.
\newblock Entropy and the consistent estimation of joint distributions.
\newblock {\em Annals of Probability}, 23:\penalty0 960--977, 1994.
\newblock See also an important Correction, \textit{Annals of Probability},
  \textbf{24} (1996): 541--545.

\bibitem[Murphy(1996)]{Murphy-passively-learning}
Kevin~P. Murphy.
\newblock Passively learning finite automata.
\newblock Technical Report 96-04-017, Santa Fe Institute, 1996.
\newblock URL
  \url{http://www.santafe.edu/sfi/publications/wpabstract/199604017}.

\bibitem[Palmer et~al.(2000)Palmer, Fairall, and
  Brewer]{Palmer-complexity-in-atmo}
A.~Jay Palmer, C.~W. Fairall, and W.~A. Brewer.
\newblock Complexity in the atmosphere.
\newblock {\em {IEEE} Transactions on Geoscience and Remote Sensing},
  38:\penalty0 2056--2063, 2000.

\bibitem[Palmer et~al.(2002)Palmer, Schneider, and
  Benjamin]{Palmer-inference-versus-imprint}
A.~Jay Palmer, T.~L. Schneider, and L.~A. Benjamin.
\newblock Inference versus imprint in climate modeling.
\newblock {\em Advances in Complex Systems}, 5:\penalty0 73--89, 2002.

\bibitem[Pearl(2000)]{Pearl-causality}
Judea Pearl.
\newblock {\em Causality: {Models}, Reasoning, and Inference}.
\newblock Cambridge University Press, Cambridge, England, 2000.

\bibitem[Perry and Binder(1999)]{Perry-Binder-finite-stat-compl}
Nicol{\'a}s Perry and P.-M. Binder.
\newblock Finite statistical complexity for sofic systems.
\newblock {\em Physical Review E}, 60:\penalty0 459--463, 1999.

\bibitem[Press et~al.(1992)Press, Teukolsky, Vetterling, and
  Flannery]{Numerical-Recipes-in-C}
William~H. Press, Saul~A. Teukolsky, William~T. Vetterling, and Brian~P.
  Flannery.
\newblock {\em Numerical Recipes in {C}: {The} Art of Scientific Computing}.
\newblock Cambridge University Press, Cambridge, England, 2nd edition, 1992.
\newblock URL \url{http://lib-www.lanl.gov/numerical/}.

\bibitem[Rayner and Best(1989)]{Rayner-Best-smooth-tests}
J.~C.~W. Rayner and D.~J. Best.
\newblock {\em Smooth Tests of Goodness of Fit}.
\newblock Oxford University Press, Oxford, 1989.

\bibitem[Rissanen(1983)]{Rissanen-1983}
Jorma Rissanen.
\newblock A universal data compression system.
\newblock {\em IEEE Transactions in Information Theory}, IT-29:\penalty0
  656--664, 1983.

\bibitem[Shalizi(2001)]{CRS-thesis}
Cosma~Rohilla Shalizi.
\newblock {\em Causal Architecture, Complexity and Self-Organization in Time
  Series and Cellular Automata}.
\newblock PhD thesis, University of Wisconsin-Madison, 2001.
\newblock URL \url{http://www.santafe.edu/$\sim$shalizi/thesis/}.

\bibitem[Shalizi and Crutchfield(2001)]{CMPPSS}
Cosma~Rohilla Shalizi and James~P. Crutchfield.
\newblock Computational mechanics: {P}attern and prediction, structure and
  simplicity.
\newblock {\em Journal of Statistical Physics}, 104:\penalty0 819--881, 2001.
\newblock E-print, arxiv.org, cond-mat/9907176.

\bibitem[Shalizi and Crutchfield(2002)]{Bottleneck-note}
Cosma~Rohilla Shalizi and James~P. Crutchfield.
\newblock Information bottlenecks, causal states, and statistical relevance
  bases: {How} to represent relevant information in memoryless transduction.
\newblock {\em Advances in Complex Systems}, 5:\penalty0 91--95, 2002.
\newblock E-print, arxiv.org, nlin.AO/0006025.

\bibitem[Spirtes et~al.(2001)Spirtes, Glymour, and
  Scheines]{Spirtes-Glymour-Scheines}
Peter Spirtes, Clark Glymour, and Richard Scheines.
\newblock {\em Causation, Prediction, and Search}.
\newblock Adaptive Computation and Machine Learning. MIT Press, Cambridge,
  Massachusetts, 2nd edition, 2001.

\bibitem[Stolcke and Omohundro(1993)]{Stolcke-Omohundro}
A.~Stolcke and S.~Omohundro.
\newblock Hidden {Markov} model induction by {Bayesian} model merging.
\newblock In Stephen~Jos{\'e} Hanson, J.~D. Gocwn, and C.~Lee Giles, editors,
  {\em Advances in Neural Information Processing Systems}, volume~5, pages
  11--18. Morgan Kaufmann, San Mateo, California, 1993.

\bibitem[Tino and Dorffner(2001)]{Tino-Dorffner-fractal}
Peter Tino and Georg Dorffner.
\newblock Predicting the future of discrete sequences from fractal
  representations of the past.
\newblock {\em Machine Learning}, 45:\penalty0 187--217, 2001.

\bibitem[Trakhtenbrot and Barzdin(1973)]{Trakhtenbrot-and-Barzdin}
B.~A. Trakhtenbrot and Ya.~M. Barzdin.
\newblock {\em Finite Automata}.
\newblock North-Holland, Amsterdam, 1973.

\bibitem[Upper(1997)]{Upper-thesis}
Daniel~R. Upper.
\newblock {\em Theory and Algorithms for Hidden {Markov} Models and Generalized
  Hidden {Markov} Models}.
\newblock PhD thesis, University of California, Berkeley, 1997.
\newblock URL \url{http://www.santafe.edu/projects/CompMech/}.

\bibitem[Vapnik(2000)]{Vapnik-nature}
Vladimir~N. Vapnik.
\newblock {\em The Nature of Statistical Learning Theory}.
\newblock Statistics for Engineering and Information Science. Springer-Verlag,
  Berlin, 2nd edition, 2000.

\bibitem[Varn et~al.(2002)Varn, Canright, and
  Crutchfield]{Varn-beyond-the-fault}
Dowman~P. Varn, Geoff~S. Canright, and James~P. Crutchfield.
\newblock Discovering planar disorder in close-packed structures from {X}-ray
  diffraction: Beyond the fault model.
\newblock {\em Physical Review B}, forthcoming, 2002.
\newblock E-print, arxiv.org, cond-mat/0203290.

\bibitem[Vidyasagar(1997)]{Vidyasagar-learning}
Mathukumalli Vidyasagar.
\newblock {\em A Theory of Learning and Generalization: {With} Applications to
  Neural Networks and Control Systems}.
\newblock Communications and Control Engineering. Springer-Verlag, Berlin,
  1997.

\bibitem[Weiss(1973)]{Weiss-1973}
Benjamin Weiss.
\newblock Subshifts of finite type and sofic systems.
\newblock {\em Monatshefte f{\"u}r Mathematik}, 77:\penalty0 462--474, 1973.

\bibitem[Willems et~al.(1995)Willems, Shtarkov, and
  Tjalkens]{Willems-Shtarkov-Tjalkens-CTW}
Frans Willems, Yuri Shtarkov, and Tjalling Tjalkens.
\newblock The context-tree weighting method: {Basic} properties.
\newblock {\em {IEEE} Transactions on Information Theory}, IT-41:\penalty0
  653--664, 1995.

\bibitem[Young and Crutchfield(1993)]{fluctuation-spectroscopy}
Karl Young and James~P. Crutchfield.
\newblock Fluctuation spectroscopy.
\newblock {\em Chaos, Solitons, and Fractals}, 4:\penalty0 5--39, 1993.

\end{thebibliography}

\end{document}